\documentclass[journal,10pt,twocolumn]{IEEETran}
\usepackage[utf8]{inputenc}

\usepackage{threeparttable}
\usepackage{cite}
\usepackage{amsmath,amssymb,amsfonts}
\usepackage{algorithmic}
\usepackage{graphicx}
\usepackage{booktabs}
\usepackage{textcomp}
\usepackage{indentfirst}
\usepackage{graphicx}
\usepackage{csquotes}
\usepackage{multirow}
\usepackage{tabularx}
\usepackage{float}
\usepackage{xcolor}
 \usepackage{url}
\usepackage{framed}
\usepackage{comment}
\usepackage{caption2}
\usepackage{algorithm2e,setspace}
\usepackage{color,colortbl}
\usepackage{hhline}
\usepackage{graphicx,subfigure}

\SetKwProg{Fn}{Function}{}{}
\definecolor{Gray}{gray}{0.9}
\newcolumntype{g}{>{\columncolor{Gray}}c}

\usepackage{bbding}

\makeatletter
\DeclareRobustCommand\onedot{\futurelet\@let@token\@onedot}
\def\@onedot{\ifx\@let@token.\else.\null\fi\xspace}

\makeatother

\renewcommand{\figurename}{Figure}
\def\tablename{Table}

\begin{document}
% \captionsetup[figure]{labelformat={default},labelsep=period,name={Figure }}
% \captionsetup{compatibility=false}

\title{Improving transferability of 3D adversarial attacks with scale and shear transformations}

\author{Jinlai Zhang, Yinpeng Dong, Jun Zhu, Jihong Zhu, Minchi Kuang, Xiaming Yuan\\

\thanks{Corresponding author: Jihong Zhu. Jinlai Zhang, Jihong Zhu are with the College of Mechanical Engineering, Guangxi University, Nanning, China (Email: cuge1995@gmail.com, jihong\_zhu@hotmail.com). Yinpeng Dong, Jun Zhu are with the Dept. of Comp. Sci. and Tech., Institute for AI, Tsinghua-Bosch Joint ML Center, THBI Lab, BNRist Center, Tsinghua University, Beijing, China; RealAI (Email: \{dongyinpeng, dcszj\}@mail.tsinghua.edu.cn). Jihong Zhu, Minchi Kuang, and Xiaming Yuan are with the Department of Precision Instrument, Tsinghua University, Beijing, China (Email: kuangmc@mail.tsinghua.edu.cn, xmyuan@tsinghua.edu.cn, jihong\_zhu@hotmail.com). This work was done when Jinlai Zhang was intern at RealAI.} } 

\maketitle

\begin{abstract}

Previous work has shown that 3D point cloud classifiers can be vulnerable to adversarial examples. However, most of the existing methods are aimed at white-box attacks, where the parameters and other information of the classifiers are known in the attack, which is unrealistic for real-world applications. In order to improve the attack performance of the black-box classifiers, the research community generally uses the transfer-based black-box attack. However, the transferability of current 3D attacks is still relatively low. To this end, this paper proposes Scale and Shear (SS) Attack to generate 3D adversarial examples with strong transferability. Specifically, we randomly scale or shear the input point cloud, so that the attack will not overfit the white-box model, thereby improving the transferability of the attack. Extensive experiments show that the SS attack proposed in this paper can be seamlessly combined with the existing state-of-the-art (SOTA) 3D point cloud attack methods to form more powerful attack methods, and the SS attack improves the transferability over 3.6 times compare to the baseline. Moreover, while substantially outperforming the baseline methods, the SS attack achieves SOTA transferability under various defenses. Our code will be available online at: {\footnotesize{\url{https://github.com/cuge1995/SS-attack}}}.
\end{abstract}


\section{Introduction}
In recent years, with the rapid development of deep learning \cite{lecun2015deepnature}, deep learning models have been deployed in numerous fields. Among them, 3D perception systems based on deep learning are increasingly deployed in safety-critical commercial environments such as autonomous driving and robotics. However, deep learning models are vulnerable to small perturbations deliberately added by attackers \cite{benchmarkingimage}. The research on how to generate adversarial examples can help us understand the robustness of different models \cite{szegedy2014intriguing,adsurvey} and further find the shortcomings of current deep learning models, which can then improve the robustness of current models.

In order to better study the robustness of current 3D point cloud classifiers, many works \cite{knnattack, advpc, aof_attack} have been proposed to generate 3D adversarial point cloud samples. Most of them are based on the CW\cite{cwattack} optimization framework.
Generally speaking, attack algorithms based on the CW optimization framework can achieve higher success rates in the white-box setting, in which the attacker has complete knowledge of the network structure and weights. However, if these adversarial examples are tested on different networks (whether in terms of network structure, weights, or both, the attack is not known), the transferability of the attack based on the CW optimization framework is relatively low. This is due to the fact that optimization-based attacks tend to overfit specific network parameters (i.e., have a high white-box success rate), so the generated adversarial examples are rarely transferable to other networks (i.e., have a low black-box success rate).
Previous work has also studied methods to improve the transferability of 3D point cloud attacks. For example, \cite{advpc} proposed the AdvPC attack method. The core idea is to jointly attack the original input point cloud and the point cloud reconstructed by an autoencoder, so that it will relies less on white-box models and better generalize to different network models. \cite{aof_attack} proposed a point cloud frequency attack method (AOF attack), where only the low-frequency part of the 3D point cloud is optimized during the attack, so as to attack the more general features of the 3D point cloud, thereby improving the transferability of the 3D point cloud adversarial samples.

Inspired by the fact that data augmentation can reduce overfitting \cite{cutmix}, this paper aims to reduce the overfitting caused by the CW optimization framework by transforming the input 3D point cloud, thereby improving the transferability of 3D adversarial point cloud samples. To this end, this paper proposes a point cloud scaling and shear attack, named \textbf{SS attack}. Specifically, in the CW optimization framework, for each iterative optimization, SS attack first sets a hyperparameter $p_{a}$, which represents the probability of whether to transform the input point cloud, and its value is between 0 and 1. The second hyperparameter is $p_{s}$, which means that if the input point cloud needs to be transformed during the attack process, it will be sheared with the probability of $p_{s}$, otherwise it will be transformed with scale deformation. Since the SS attack only scales and shears the input point cloud, it can be seamlessly integrated with the existing attack methods. This paper compares the proposed SS attack with the current state-of-the-art methods and SS attack combination of attack methods such as 3D-Adv\cite{generatingadpoint}, kNN attack\cite{knnattack}, AdvPC attack\cite{advpc} and AOF attack\cite{aof_attack}. In each iteration of optimization, unlike traditional methods that directly maximize the loss function, we apply a transformation (e.g., random scaling, random shear) to the input point cloud with probability $p_{a}$ and maximize the loss function. Extensive experiments are performed to validate the transferable performance of SS attacks on several classical point cloud classification networks in both white-box and black-box settings. Compared with the attacking algorithms of the traditional CW optimization framework, the experimental results on ModelNet40 show that SS attack achieves significantly higher transferability in the black-box model and maintains a similar attack success rate in the white-box model. We hoped SS attack can serve as a strong baseline to evaluate the robustness of 3D point cloud classifiers and the effectiveness of different defense methods in the future.

\section{Related Work}
\textbf{Deep learning on point clouds.} 
After the poineering work of PointNet\cite{pointnet}, learning 3D representations from point cloud via deep learning has attracted much attention. The PointNet cannot learn the local features well, thus the subsequent research proposed various
deep neural networks to learn more representative features across local and global. The most famous networks are PointNet++\cite{pointnet++}, PointConv\cite{pointconv} and DGCNN\cite{dgcnn}.  The PointNet++ first randomly select a point in the input point set as the center point, and then further select the 3D points around the overlapping local area of the limited Euclidean distance around the selected center point to form a sub-area to learn the local features. PointConv learn 3D representations using the convolution paradigm,  and the convolution kernel is regarded as a nonlinear function composed of weight and density functions acting on the local 3D point coordinate system. DGCNN proposes edge convolution (EdgeConv), which continuously recalculates the neighbors of each point in the feature space of each layer. DGCNN dynamically generates a graph from the nearest point features, and then learns the 3D shape representation from the graph.

More recently, inspired by the success in natural language processing, the transformer architecture is explored in point cloud learning. The PCT\cite{guo2021pct} is a new transformer-based point cloud learning framework, compared with the original self-attention module in transformer, they propose an improved self-attention with implicit Laplacian operator and normalization. The DPCT\cite{han2022dual}  learn 3D representations by the well-designed point-wise and channel-wise multi-head self-attention simultaneously, and can semantically capture richer contextual dependencies from both location and channel perspectives. The Patchformer\cite{zhang2022patchformer} uses the Patch ATtention (PAT) and a lightweight Multi-Scale aTtention
(MST) block to learn 3D representations.

\textbf{Adversarial attacks on point clouds.} The adversarial attack on point cloud can be roughly divide into white-box attack and black-box attack. 
The white-box attack means that the attacker knows all information about the deep learning models, such as the model's architecture and the gradient of input data. 3D-Adv~\cite{generatingadpoint} generate adversarial point clouds examples by point perturbation and adding additional points. However, the adversarial point clouds examples craft by  3D-Adv have a lot outliers. Therefore, the subsequent research proposed various methods to generate adversarial examples that are less outlier point, but are still capable of attacking 3D point cloud classification models. KNN attack ~\cite{knnattack} employs kNN distance constraints to generate smoother and less perceptible adversarial point cloud examples. The Geometry-Aware Adversarial Attack ($GeoA^3$) ~\cite{geoa3} uses the geometry-aware objectives to further improved the imperceptible to humans. Moreover, the point drop attack ~\cite{pointcloudsaliencymaps} was developed by a gradient based saliency map, which iteratively remove the points with highest saliency score. 

The black-box attack can be further classified as transfer-based black-box attacks, score-based black-box attacks and decision-based black-box attacks\cite{benchmarkingimage}. For point cloud, most works focus on transfer-based black-box attack. The AdvPC \cite{advpc} uses the point cloud auto-encoder to improved the transferability of adversarial point cloud examples, and the AOF \cite{aof_attack} only perturb the low-frequency part of the 3D point cloud, so as to attack the general features of the 3D point cloud and improve the transferability of the adversarial point cloud samples.

\section{Methodology}
\subsection{An overview of classical point cloud attack}
Let a 3D point cloud as $\mathcal{X} \in \mathbb{R}^{N \times 3}$, where each line is the 3D coordinates of a point. Usually, for the classification model $\mathcal{F}:\mathcal{X} \rightarrow \mathcal{Y}$, it mapping the point cloud to its corresponding class label, and the adversarial point cloud sample can be obtained by within $l_{p}$ a sphere of size $\epsilon_p$, where $p$ can be $1$, $2$, and $\infty$. The adversarial point cloud can be expressed as $\mathcal{X}' = \mathcal{X} + \Delta$, where $\Delta$ is the added small perturbation.

\begin{figure*}[!tp]
    \centering
    \includegraphics[width=\textwidth]{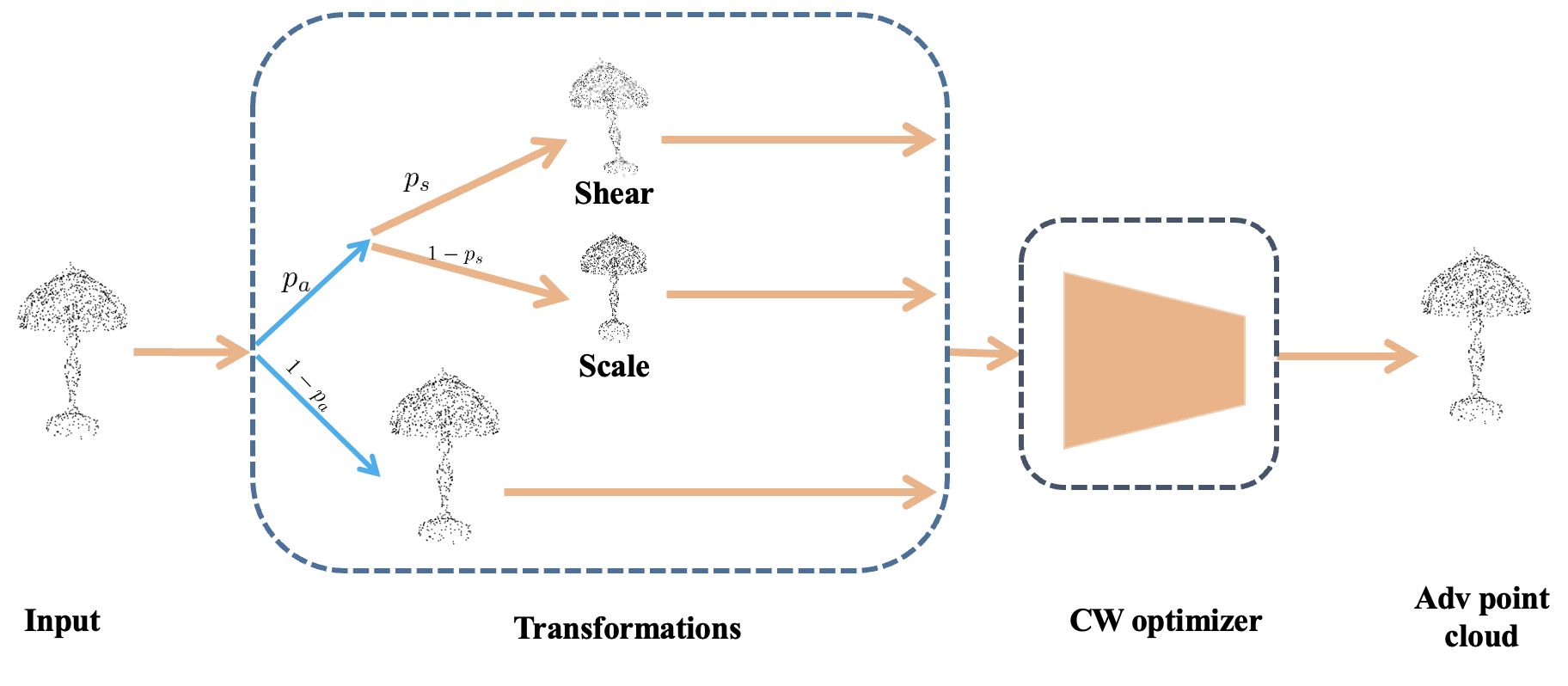}
    \caption{The pipeline of SS attack.}
    \label{pl_compare}
\end{figure*}

Most of the previous 3D point cloud attack methods crafted the 3D adversarial point cloud sample $\mathcal{X}'\in\mathbb{R}^{N \times 3}$ by solving the following optimization problem:
\begin{equation}
     \min_{\Delta} l(\mathcal{X}, \mathcal{X}'), \quad s.t.D(\mathcal{X}, \mathcal{X'})\le\epsilon
\end{equation}
where the $l()$ loss function generally includes two items, which are the adversarial loss function $l_{adv}$ and the distance loss function $l_{dis}$, the $l_{adv}$ is defined as following:
\begin{equation}
\label{eqn:margin loss}
     l_{adv}(\mathcal{X}') = \max(\mathbf{Z}(\mathcal{X}')_{y_{gt}} - \max_{y \neq y_{gt}}(\mathbf{Z}(\mathcal{X}')_y) + \kappa, 0)
\end{equation}
where $y_{gt}$ is the ground truth class, $y$ is any class not equal to $y_{gt}$, $\mathbf{Z(\cdot)}$ is the output of the logits layer, and $\kappa$ is loss margin. Therefore, the previous 3D point cloud attack methods can be summarized as:

\textbf{3D-Adv\cite{generatingadpoint}} crafts 3D adversarial point cloud samples by optimizing the following loss function:
\begin{equation}
l_{3d-adv} = l_{adv}(\mathcal{X}') + \|\mathcal{X}'-\mathcal{X}\|_{2}^{2}
\end{equation}

\textbf{kNN attack\cite{knnattack}} crafts 3D adversarial point cloud samples by optimizing the following loss function:
\begin{equation}
l_{knn_{adv}} = l_{adv}(\mathcal{X}') + l_{knn}(\mathcal{X}')
\end{equation}
where $l_{knn}$ expands to:
\begin{equation}
l_{knn}(\mathcal{X}') = \frac{1}{\|\mathcal{X}'\|} \sum_{p \in \mathcal{X}'} w_{p} \cdot d_ {p}
\end{equation}
where $ d_{p}=\frac{1}{k}\left(\sum_{p^{\prime} \in \mathrm{kNN}(p, \mathcal{X}')}\left\|p-p ^{\prime}\right\|_{2}^{2}\right)$, for point $p \in \mathcal{X}$, $\mathrm{kNN}(p, \mathcal{X}' )$ represents its nearest $k$ nearest neighbors in Euclidean space, if $d_{p}$ is greater than the defined threshold, then $w_{p} = 1$, otherwise $w_{p} = 0$.

\textbf{AdvPC attack\cite{advpc}} crafts 3D adversarial point cloud samples by optimizing the following loss function:
  \begin{equation}
  \label{eqn:aof loss}
      l_{advpc} = \left(1-\gamma\right)l_{adv}(\mathcal{X}')+\gamma l_{adv}(\mathcal{X}_{encoder}'))
\end{equation}
where $\mathcal{X}_{encoder}'$ is the 3D point cloud reconstructed by the auto-encoder, and $\gamma$ is the hyperparameter of AdvPC attack.

\textbf{AOF attack\cite{aof_attack}} crafts 3D adversarial point cloud samples by optimizing the following loss function:
\begin{equation}
\label{eqn:aof loss}
      l_{aof} = \left(1-\gamma\right)l_{adv}(\mathcal{X}')+\gamma l_{adv}(\mathcal{X}_{lfc}'))
\end{equation}
where $\mathcal{X}_{lfc}'$ represents the the low frequency part of point cloud, and $\gamma$ is the hyperparameter of AOF attack.

\subsection{Scale and shear attack (SS attack)}
Dong et al. \cite{dong2018boosting} claimed that the generation of adversarial samples is similar to the training of deep neural models, and the transferability of adversarial samples can be analogized to the generalization of training models. In this view, transforming the input can be viewed as data augmentation. In the 2D image domain, various input transformations have been proposed \cite{xie2019improving,dong2019evading} that can improve adversarial transferability. However, in the current 3D point cloud attack methods, there is a lack of corresponding research. To fill this research gap, in this paper, a point cloud scale and shear attack method is proposed to alleviate the attack overfitting problem brought by the CW optimization framework.

The overall framework of the point cloud scale and shear attack proposed in this paper is shown in the figure \ref{pl_compare}. For each iterative in optimization, SS attack first sets a hyperparameter $p_{a}$, which represents the probability of whether to transform the input point cloud, its value is between 0 and 1. The second hyperparameter is $p_{s}$, which means that if the input point cloud needs to be transformed during the attack process, it will be sheared with the probability of $p_{s}$, otherwise it will be transformed with scale deformation.

\textbf{SS-3D-Adv attack.} We first proposes the SS-3D-Adv attack method, which performs scale and shear transformation on the input of the point cloud of each iteration, and crafts 3D adversarial point cloud samples by optimizing the following loss function:
\begin{equation}
l_{ss-3d-adv} = l_{adv}(T(\mathcal{X}')) + \|\mathcal{X}'-\mathcal{X}\|_{2}^{2}
\end{equation}
where $T(\mathcal{X}')$ can be further expressed as:
\begin{equation}
T\left(\mathcal{X}' ; p_{a}, p_{s}\right)= \begin{cases}T_{scale}\left(\mathcal{X}'\right) & \text { with probability } p_{a}*p_{s} \\ T_{shear}\left(\mathcal{X}'\right) & \text { with probability } p_{a}*(1-p_{s}) \\ \mathcal{X}' & \text { with probability } 1-p_{a}\end{cases}
\label{pl_transform}
\end{equation}
where $T_{scale}()$ represents the scale transformation, and $T_{shear}()$ represents the oblique transformation.

\textbf{SS-kNN attack.} Similarly, the SS attack proposed in this paper can also easily combine kNN attacks to form an SS-kNN attack method, which crafts 3D adversarial point cloud samples by optimizing the following loss function:
\begin{equation}
l_{ss-knn_{adv}} = l_{adv}(T(\mathcal{X}')) + l_{knn}(T(\mathcal{X}'))
\end{equation}

\textbf{SS-AdvPC attack.} For the AdvPC attack, we only scales and shears the point cloud before passing the autoencoder, so as to combine into a new SS-AdvPC attack method, which crafts 3D adversarial point cloud samples by optimizing the following loss function:
  \begin{equation}
  \label{eqn:aof loss}
      l_{ss-advpc} = \left(1-\gamma\right)l_{adv}(T(\mathcal{X}'))+\gamma l_{adv}(\mathcal{X}_{encoder}' ))
\end{equation}

\textbf{SS-AOF attack.} For AOF attack, we only scales and shears the input point cloud, and keeps the low-frequency part of the 3D point cloud unchanged, which is combined into a new SS-AOF attack method. The following loss function is used to crafts 3D adversarial point cloud samples:
\begin{equation}
\label{eqn:aof loss}
      l_{ss-aof} = \left(1-\gamma\right)l_{adv}(T(\mathcal{X}'))+\gamma l_{adv}(\mathcal{X}_{lfc}' ))
\end{equation}

The transformation $T()$ involved in all the above attack methods is shown in the formula \ref{pl_transform}. It can be seen from the above description that the method proposed in this paper has the characteristics of plug-and-play and can be integrated with the current mainstream 3D point cloud attack methods.

\subsection{Point cloud deformation analysis}\label{unified}
\begin{figure}[htp]
    \centering
    \includegraphics[width=\linewidth]{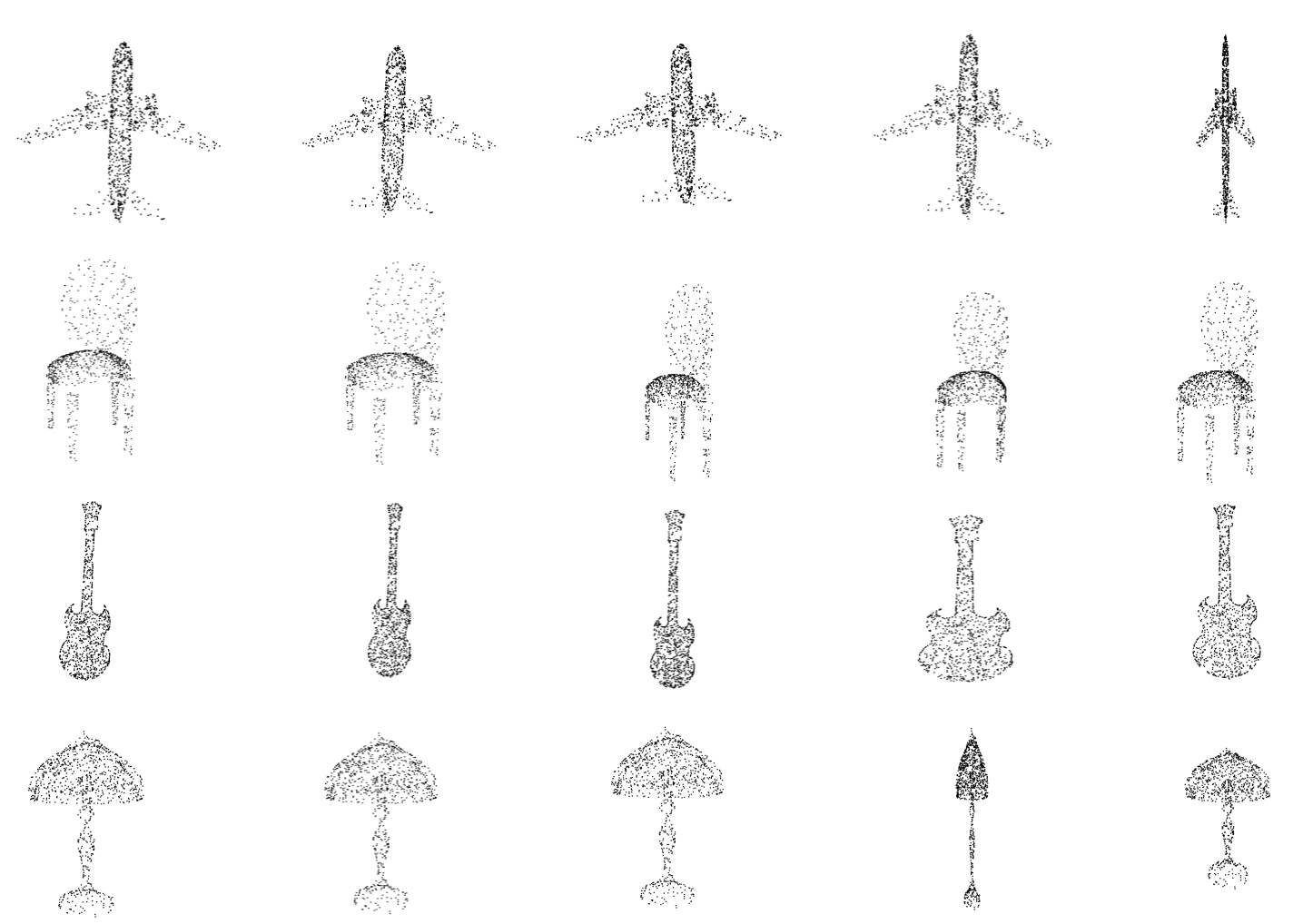}
    \caption{Visualization of different scale intensity.}
    \label{scale_compare}
\end{figure}

We represents the 3D point cloud before deformation as $\mathcal{X} \in \mathbb{R}^{N \times 3}$, and the 3D point cloud after deformation as $\mathcal {X}_t \in \mathbb{R}^{N \times 3}$, where each line is the three-dimensional coordinates of a point. Then the scale deformation of the 3D point cloud can be expressed as:
\begin{equation}
\mathcal{X}_t = \mathcal{X}\begin{bmatrix}
a & b & c\\
\end{bmatrix}
\end{equation}
where $a,b,c$ represent the scale intensities of the three-dimensional point cloud $x,y,z$ axes, and the scale intensities in this section are taken from random numbers within a certain range. Figure \ref{scale_compare} shows the visualization of 3D point cloud with different degrees of scale  deformation, from left to right: 3D point cloud without deformation, scale intensity range $[0.9, 1.1]$, $[0.5, 1.5]$, $[0.4, 2.5]$, $[0.2, 5]$. It can be seen that when the range is larger, the deformation of the object is more serious.

\begin{figure}[!thp]
    \centering
    \includegraphics[width=\linewidth]{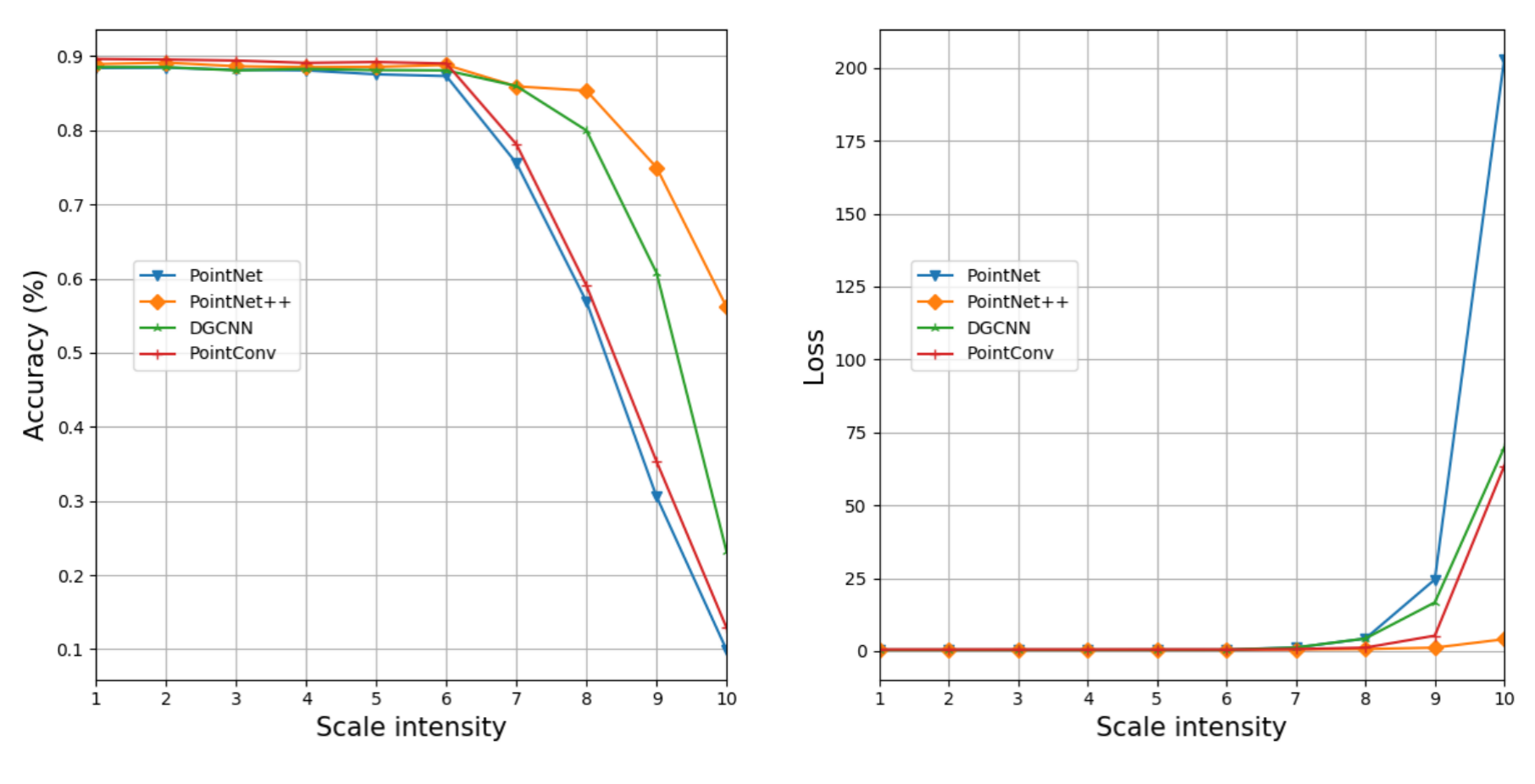}
    \caption{The accuracy and loss of different scale intensities.}
    \label{scale_inten}
\end{figure}

In order to further clarify the impact of these deformations on the 3D point cloud classifier, we selects 500 point cloud samples in the ModelNet40 test set, and visualizes the effects of different scale intensity on the accuracy and loss of the 3D point cloud classifier. As shown in Figure \ref{scale_inten}, the scale deformation intensity from left to right are: no deformation, scaled with range in $[0.9, 1.1]$, $[0.8, 1.25]$, $[0.6, 1.428]$, $[ 0.5, 1.5]$, $[0.4, 2.5]$, $[0.3, 3.33]$, $[0.2, 5]$. As can be seen from the figure, when the degree of scale deformation is between $[0.5, 1.5]$, for PointNet, PointNet++, PointConv and DGCNN, the model accuracy and loss function have little effect. Therefore, the degree of scale deformation is selected between $[0.5, 1.5]$.

\begin{figure}[htp]
    \centering
    \includegraphics[width=\linewidth]{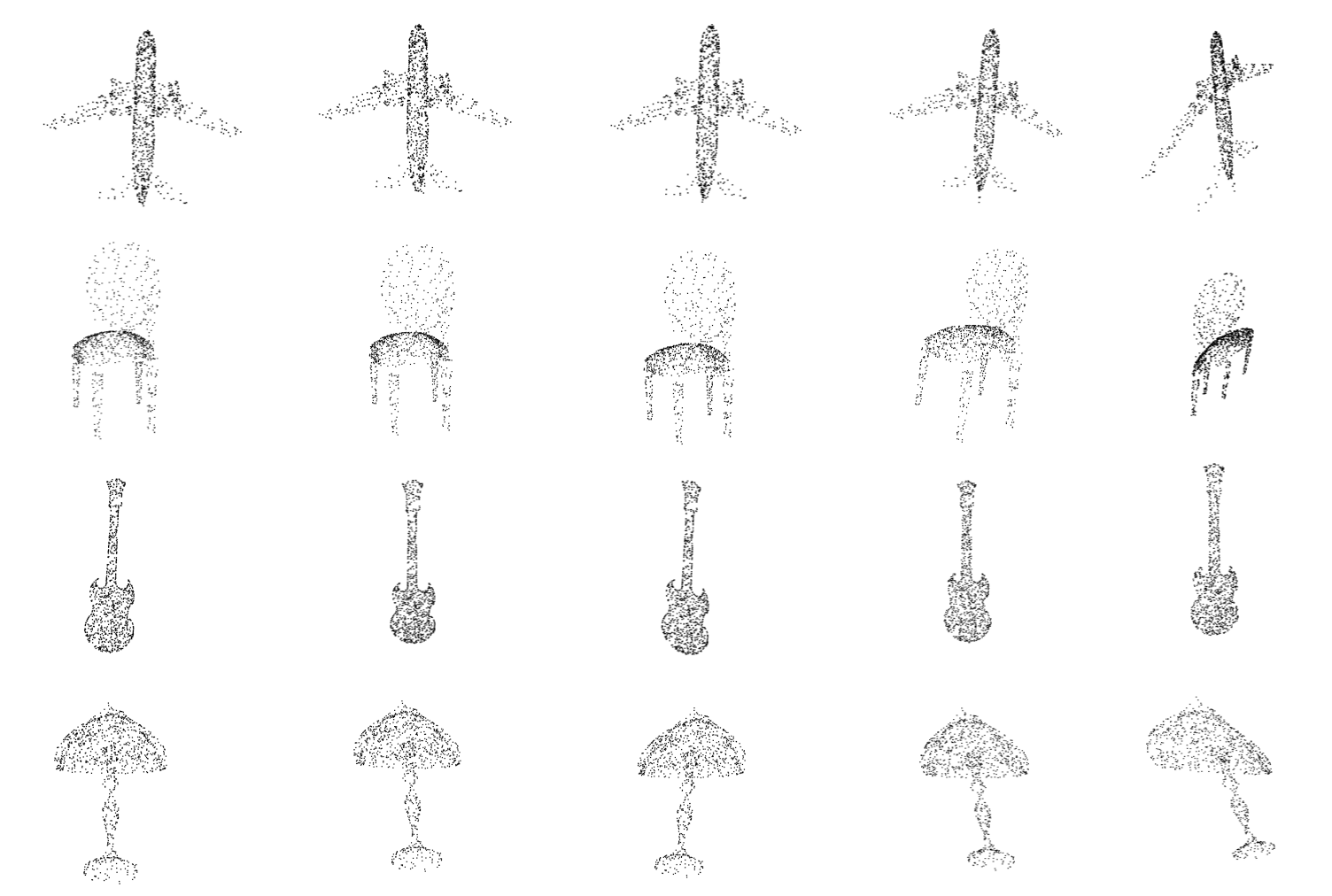}
    \caption{Visualization of different shear intensity.}
    \label{shrear_compare}
\end{figure}

Similarly, the 3D point cloud shear deformation can be expressed as:
\begin{equation}
\mathcal{X}_t = \mathcal{X}\begin{bmatrix}
1 & 0 & d\\
e & 1 & f\\
f & 0 & 1\\
\end{bmatrix}
\end{equation}
where $d, e, f, g$ represent the shear deformation intensity of the 3D point cloud respectively, and its positive and negative values are also random. Figure \ref{shrear_compare} shows the visualization of the 3D point cloud with different degrees of shear deformation, from left to right are: 3D point cloud without deformation, shear deformation intensity range are $[0, 0.1]$, $[0.05, 0.15]$, $[0.15, 0.25]$, $ [0.35, 0.45]$. It can be seen that when the range of shear deformation intensity is larger, the deformation of the object is more serious.

\begin{figure}[!htp]
    \centering
    \includegraphics[width=\linewidth]{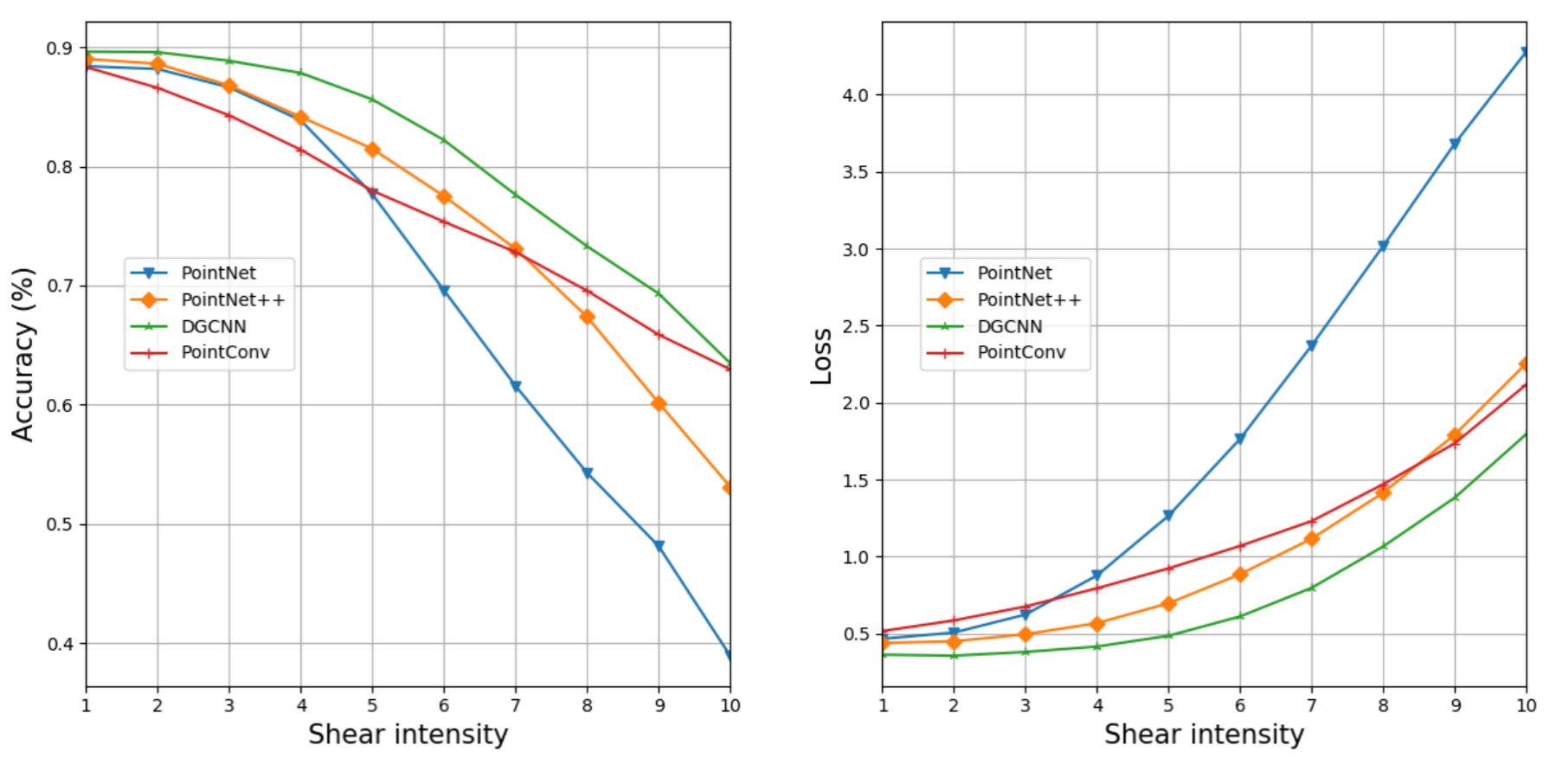}
    \caption{The accuracy and loss of different shear intensities.}
    \label{shrear_inten}
\end{figure}

In order to further clarify the influence of these deformations on the 3D point cloud classifier, we selects 500 point cloud samples in the ModelNet40 test set, and visualizes the influence of different shear deformation intensities on the accuracy and loss function of the 3D point cloud classifier. As shown in the figure \ref{shrear_inten}, the intensity of shear deformation from left to right is: no deformation, shear deformation in rnage of $[0, 0.1]$, $[0.05, 0.15]$, $[0.1, 0.2]$, $ [0.2, 0.3]$, $[0.25, 0.35]$, $[0.3, 0.4]$, $[0.35, 0.45]$, $[0.40, 0.50]$, $[0.45, 0.55]$. It can be seen from the figure that when the degree of shear deformation is between $[0, 0.15]$, for PointNet, PointNet++, PointConv and DGCNN, the model accuracy and loss function have little effect. Therefore, the shear deformation intensity is selected between $[0, 0.15]$.

\subsection{Relationship between various 3D point cloud attack methods}

\begin{figure}[htp]
    \centering
    \includegraphics[width=\linewidth]{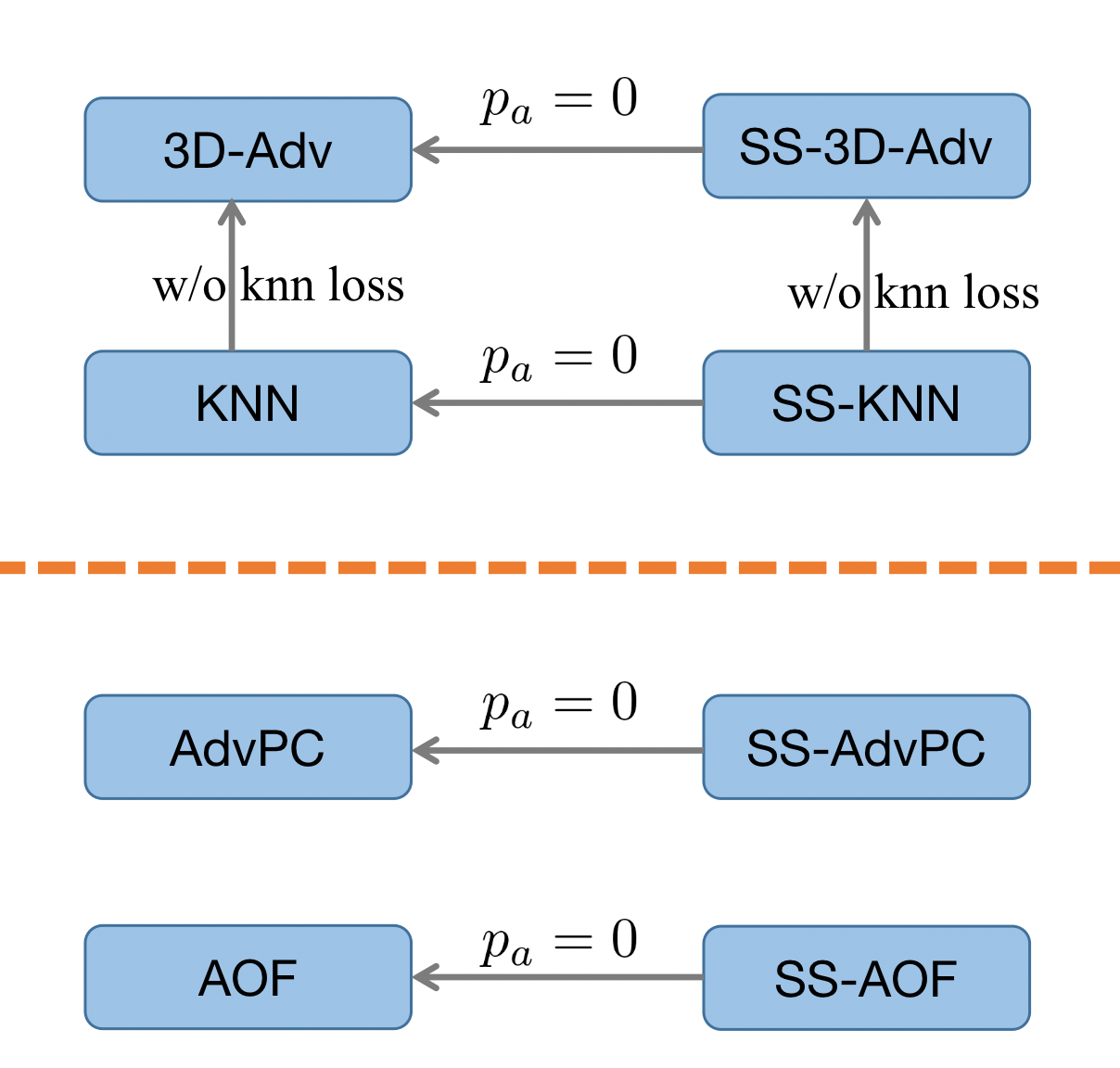}
    \caption{The relationship between various 3D point cloud attack methods.}
    \label{att_relations}
\end{figure}

The attacks mentioned above all belong to the optimization-based attack methods, and they can be correlated by different parameter settings. In order to better understand the relationship between various 3D point cloud attack methods, this paper gives a brief overview of various 3D point cloud attack methods, as shown in Figure \ref{att_relations}. Generally,
if the transformation probability $p_a = 0$, the SS-3D-Adv attack method degenerates into a 3D-Adv attack method, the SS-3D-Adv attack method degenerates into a 3D-Adv attack method, and the SS-KNN attack method degenerates into a 3D-KNN attack method method, the SS-AdvPC attack method degenerates into the AdvPC attack method, and the SS-AOF attack method degenerates into the AOF attack method.

Moreover, if SS-KNN does not have a loss term of $l_{knn}()$, the SS-KNN attack method degenerates into the SS-3D-Adv attack method, and the KNN attack method degenerates into the 3D-ADv attack method.

\section{Experiments}

\subsection{Experimental Setup}
\textbf{Dataset. } Following the previous state-of-the-art 3D adversarial point cloud attack algorithm \cite{aof_attack,geoa3}, the experiments in this paper are performed on the ModelNet40\cite{modelnet40} dataset. The ModelNet40 dataset contains synthetic objects, released by the Graphics Laboratory of Princeton University, USA. As the most widely used benchmark dataset for point cloud analysis, ModelNet40 is popular due to its diverse categories, clear shapes, and well-structured datasets. ModelNet40 consists of 40 categories (such as airplanes, cars, plants, lights), of which 9843 are used for training and the remaining 2468 are used for testing. The corresponding point cloud data are uniformly sampled from the 3D mesh surface.

\textbf{Baseline model and attack method. }
For a fair comparison with previous work, this paper adopts the same deep learning model as AOF\cite{aof_attack}, namely: PointNet\cite{pointnet}, pioneering work in the field of 3D point cloud deep learning; PointNet++\cite {pointnet++}, which learn local and global features at the same time; DGCNN\cite{dgcnn}, which model 3D point cloud into a graph; PointConv\cite{pointconv}, which introduce convolution into point cloud classification network. We use the pretrained model kindly provided by AOF\cite{aof_attack} attack. In addition, the classic 3D point cloud attack methods are also used for comparison, they are: 3D-Adv\cite{generatingadpoint}, the first 3D point cloud attack method; kNN attack\cite {knnattack}, a 3D point cloud attack method that can generate less surface disturbance of point clouds; AdvPC attack\cite{advpc}, a 3D point cloud attack method that uses a 3D point cloud autoencoder to effectively improve the transferability of adversarial point cloud samples; AOF attack\cite {aof_attack}, the best 3D point cloud attack method with highest transferability, the key idea of it is optimized to attack the more general features of 3D point cloud, thereby improving the transferability of 3D adversarial point cloud samples.

\textbf{Experimental details. }
The experiments in this paper were performed on a Linux server with five NVIDIA 3090 GPUs, each with 24GB memory. The operating system is Ubuntu 20.04. All deep learning models are built with the pytorch\cite{paszke2019pytorch} framework. Similar to the AOF\cite{aof_attack} attack, this paper uses Adam\cite{kingma2014adam} as the optimizer for various attack methods. For the 3D-Adv\cite{generatingadpoint} attack, the number of iterations is 500, the number of binary search steps is 10, and the learning rate of the Adam optimizer is set to 0.01, which is consistent with its original paper. For the kNN\cite{knnattack} attack, the number of iterations is 2500 steps, the minimum margin in its adversarial loss function is set to 15, and the learning rate of the Adam optimizer is set to 0.001. For the AdvPC\cite{advpc} attack, the number of iterations is 200, the number of binary search steps is 2, and the learning rate of the Adam optimizer is set to 0.01. In addition, due to the autoencoder used in the AdvPC attack, this paper adopts the open sourced the pretrained model of the AdvPC autoencoder by AOF attack \cite{aof_attack}. For the AOF attack, the number of iterations is 200, the number of binary search steps is 2, and the learning rate of the Adam optimizer is set to 0.01, which is consistent with its original paper data. In order to further reduce the overfitting of the white-box model, the SS attacks in this paper do not use binary search for further optimization. The rest of the hyperparameter settings are the same as the above attack methods. Moreover, we set $p_a = 0.5$, $p_s = 0.5$ for SS-3D-Adv, while $p_a = 0.7$, $p_s = 0.7$ for SS-kNN, SS-AdvPC, and SS-AOF.

\textbf{Evaluation metrics. } As emphasized in the AOF attack \cite{aof_attack}, the accuracy of most 3D point cloud classifiers on the test set is basically below 94\%, only using the classification accuracy to evaluate transferability will bring a certain bias. To this end, this paper adopts the same evaluation metrics as AOF attacks\cite{aof_attack}, namely:

\begin{equation}
\label{eqn:trans}
    T_{rans} = \frac{\left|S_{v2t}^{tm}\right|}{\left|S^{tm}\right|} * 100\%
\end{equation}
where $S^{tm}$ denotes the clean samples correctly classified by the transfer model, and the corresponding adversarial samples generated by the victim model of $S^{tm}$ as $S_{adv }^{tm }$. In the set $S_{adv}^{tm}$, we denote the samples misclassified by the transfer model as the set $S_{v2t}^{tm}$.

\subsection{The transferability of untargeted attack.}

In this section, extensive experiments were performed to validate whether the SS attack is more transferable to different types of target 3D point cloud classifiers.
Experiments are performed on the ModelNet40 dataset. 3D adversarial point cloud samples generated by SS-3D-Adv, SS-kNN, SS-AdvPC, SS-AOF, and 3D-Adv, kNN, AdvPC, and AOF are evaluated. The target networks include Pointnet, Pointnet++, DGCNN, and PointConv.
We implement the untargeted attack, that is, after adding a small perturbation to the original point cloud, the target point cloud classifier is not successful to classifying the adversarial point cloud. 

As shown in Table \ref{p1untransfer}, the column name as the same as victim model represents the white-box attack, and the other columns represent the black-box attack. When the attacked white-box model is PointNet, the DGCNN, PointNet++, and PointConv are black-box models for adversarial sample transfer for first 8 rows. When PointNet is used as the victim model, the SS attack has a significant improvement of transferability compared to the baseline method. The attack method combined with SS attack outperforms all other attack methods without combined SS attack by a large margin on all black-box models and maintains a high success rate on all white-box models. It can be seen that SS-KNN has a large improvement compared to KNN, and the transferability is improved by a maximum of 15.3\%. For the adversarial point cloud samples generated on PointNet, the transferability of the SS-KNN attack method on PointNet++ is 26.3\%, the transferability on DGCNN is 21.7\%, and the transferability on PointConv is 6.19\%, while the KNN attack without combined SS attack only obtains 11.0\%, 3.83\% and 10.02\% transferability, respectively. For the AOF attack method with the strongest transferability, the SS-AOF attack method improves the transferability by 1.8\% on PointNet++, 1.3\% on DGCNN, and 2.0\% on PointConv. This strongly demonstrates the effectiveness of the combination of SS attack and current mainstream 3D point cloud attack methods for improving the transferability of 3D adversarial point cloud examples. When the PointNet++, PointConv and DGCNN as white-box model, we can observe similar phenomenon, which is SS attack can improve the transferability significantly on all black-box models and maintains a high attack success rate on all white-box models. 

\begin{table*}[ht]
\centering
\caption{\textbf{The transferability of untargeted attack.} We use $T_{rans}$ defined in Eq (\ref{eqn:trans}) to evaluate the transferability. 
}
\setlength{\tabcolsep}{1.7mm} %
\renewcommand{\arraystretch}{1.1} %
\begin{tabular}{cc|cccc} 
\toprule
%  &  &\multicolumn{4}{c}{$\epsilon_\infty = 0.18$}\\
Victim model & Attacks  & PointNet & PointNet++ & PointConv & DGCNN   \\
\midrule
\multirow{8}{*}{PointNet} & 3D-Adv  &  100 &  5.26 & 2.32 &  3.76\\
 & SS-3D-Adv  &  94.9 &  \bf10.7 \textcolor{green!40!gray}{\small $\uparrow$5.4} & \bf3.96 \textcolor{green!40!gray}{\small $\uparrow$1.62} &  \bf7.70 \textcolor{green!40!gray}{\small $\uparrow$3.94}\\
%  \midrule
\cline{2-6}
& KNN  & 100 &  11.0 & 3.83 &  10.2  \\
 & SS-KNN  & 98.5 &  \bf26.3 \textcolor{green!40!gray}{\small $\uparrow$15.3} & \bf6.19 \textcolor{green!40!gray}{\small $\uparrow$2.36}&  \bf21.7 \textcolor{green!40!gray}{\small $\uparrow$11.5} \\
%  \midrule
\cline{2-6}
& AdvPC  &  100 & 30.4 & 13.2 & 14.6  \\
&  SS-AdvPC  &  98.5 & \bf35.7 \textcolor{green!40!gray}{\small $\uparrow$4.3} & \bf14.5 \textcolor{green!40!gray}{\small $\uparrow$1.3} & \bf16.5 \textcolor{green!40!gray}{\small $\uparrow$1.9} \\
%   \midrule
  \cline{2-6}
&  AOF  &  99.9 & 56.8 & 35.9 & 28.7 \\
&   SS-AOF  &  99.9 & \bf58.6 \textcolor{green!40!gray}{\small $\uparrow$1.8}& \bf37.9 \textcolor{green!40!gray}{\small $\uparrow$2.0}& \bf30.0 \textcolor{green!40!gray}{\small $\uparrow$1.3}\\
 \midrule
%   \cline{2-6}
  \multirow{8}{*}{PointNet++} &3D-Adv  &  2.09 &  78.5 & 4.95 &  6.27\\
 & SS-3D-Adv  &  \bf3.26 \textcolor{green!40!gray}{\small $\uparrow$1.17}&  84.0 & \bf5.71 \textcolor{green!40!gray}{\small $\uparrow$0.76}&  \bf10.3 \textcolor{green!40!gray}{\small $\uparrow$4.03}\\
  \cline{2-6}
&KNN  & 2.54 &  99.9 & 5.57 &  6.94  \\
  &SS-KNN  & \bf5.35 \textcolor{green!40!gray}{\small $\uparrow$2.81}&  99.9 & \bf7.14 \textcolor{green!40!gray}{\small $\uparrow$1.57}&  \bf11.9 \textcolor{green!40!gray}{\small $\uparrow$4.97} \\
  \cline{2-6}
  &AdvPC  &   5.08 & 99.5 & 27.5 & 17.5  \\
  &SS-AdvPC  &  \bf5.98 \textcolor{green!40!gray}{\small $\uparrow$0.9}& 98.3 & \bf31.7 \textcolor{green!40!gray}{\small $\uparrow$4.2}& \bf21.7 \textcolor{green!40!gray}{\small $\uparrow$4.2} \\
  \cline{2-6}
   &AOF  &  5.39 & 97.3 & 41.4 & 26.4 \\
   &SS-AOF  &  \bf7.57 \textcolor{green!40!gray}{\small $\uparrow$2.18}& 97.2 & \bf46.1 \textcolor{green!40!gray}{\small $\uparrow$4.7}& \bf31.1 \textcolor{green!40!gray}{\small $\uparrow$4.7}\\
\midrule

\multirow{8}{*}{PointConv} &3D-Adv  &  1.81 &  7.87 & 84.3 & 4.47\\
 &SS-3D-Adv  &  \bf3.49 \textcolor{green!40!gray}{\small $\uparrow$1.68}&  \bf10.5 \textcolor{green!40!gray}{\small $\uparrow$2.6}& 82.7 &  \bf7.39 \textcolor{green!40!gray}{\small $\uparrow$2.92}\\
 \cline{2-6}
 &KNN  & 3.49 &  17.1 & 100 &  11.3  \\
 &SS-KNN  & \bf6.29 \textcolor{green!40!gray}{\small $\uparrow$2.80}&  \bf27.1 \textcolor{green!40!gray}{\small $\uparrow$10.0}& 100&  \bf17.4 \textcolor{green!40!gray}{\small $\uparrow$6.1}\\
 \cline{2-6}
 &AdvPC  &   4.62 & 35.3 & 99.4 & 17.7  \\
 & SS-AdvPC  &  \bf5.49 \textcolor{green!40!gray}{\small $\uparrow$0.87} & \bf42.8 \textcolor{green!40!gray}{\small $\uparrow$7.5}& 98.5 & \bf21.0 \textcolor{green!40!gray}{\small $\uparrow$3.3}\\
\cline{2-6}
 & AOF  &  4.62 & 35.8 & 97.4 & 18.2 \\
 & SS-AOF  &  \bf5.71 \textcolor{green!40!gray}{\small $\uparrow$1.09}& \bf40.3 \textcolor{green!40!gray}{\small $\uparrow$4.5}& 96.6 & \bf20.2 \textcolor{green!40!gray}{\small $\uparrow$2.0}\\

\midrule

 \multirow{8}{*}{DGCNN} &3D-Adv  &  0.99 &  5.84 & 4.76 & 100\\
  &SS-3D-Adv  &  \bf4.26 \textcolor{green!40!gray}{\small $\uparrow$3.27}&  \bf27.3 \textcolor{green!40!gray}{\small $\uparrow$21.4} & \bf14.6 \textcolor{green!40!gray}{\small $\uparrow$9.8}&  92.9\\
 \cline{2-6}
  &KNN  & 5.39 &  30.6 & 20.9 &  100  \\
  &SS-KNN  & \bf8.98 \textcolor{green!40!gray}{\small $\uparrow$3.59}&  \bf46.9 \textcolor{green!40!gray}{\small $\uparrow$16.7}& \bf31.1 \textcolor{green!40!gray}{\small $\uparrow$10.2}&  100 \\
 \cline{2-6}
  &AdvPC  &   6.94 & 60.1 & 43.5 & 93.7  \\
  &SS-AdvPC  &  \bf9.48  \textcolor{green!40!gray}{\small $\uparrow$2.54}& \bf65.8 \textcolor{green!40!gray}{\small $\uparrow$5.7}& \bf48.0 \textcolor{green!40!gray}{\small $\uparrow$4.5}& 91.5 \\
  \cline{2-6}
   &AOF  &  12.6 & 61.7 & 54.6 & 100 \\
   &SS-AOF  &  \bf13.9 \textcolor{green!40!gray}{\small $\uparrow$1.3}& \bf68.2 \textcolor{green!40!gray}{\small $\uparrow$6.7} & \bf60.0 \textcolor{green!40!gray}{\small $\uparrow$5.4}& 100\\

 \bottomrule
\end{tabular}
\label{p1untransfer}
\end{table*}

In order to further analyze the reasons for the improved transferability of the SS series of attacks, following, we visualize the 3D adversarial point cloud samples generated by SS-3D-Adv, SS-kNN, SS-AdvPC, SS-AOF, and 3D-Adv, kNN, AdvPC, and AOF attack methods.

\begin{figure}[!thp]
    \centering
    \includegraphics[width=\linewidth]{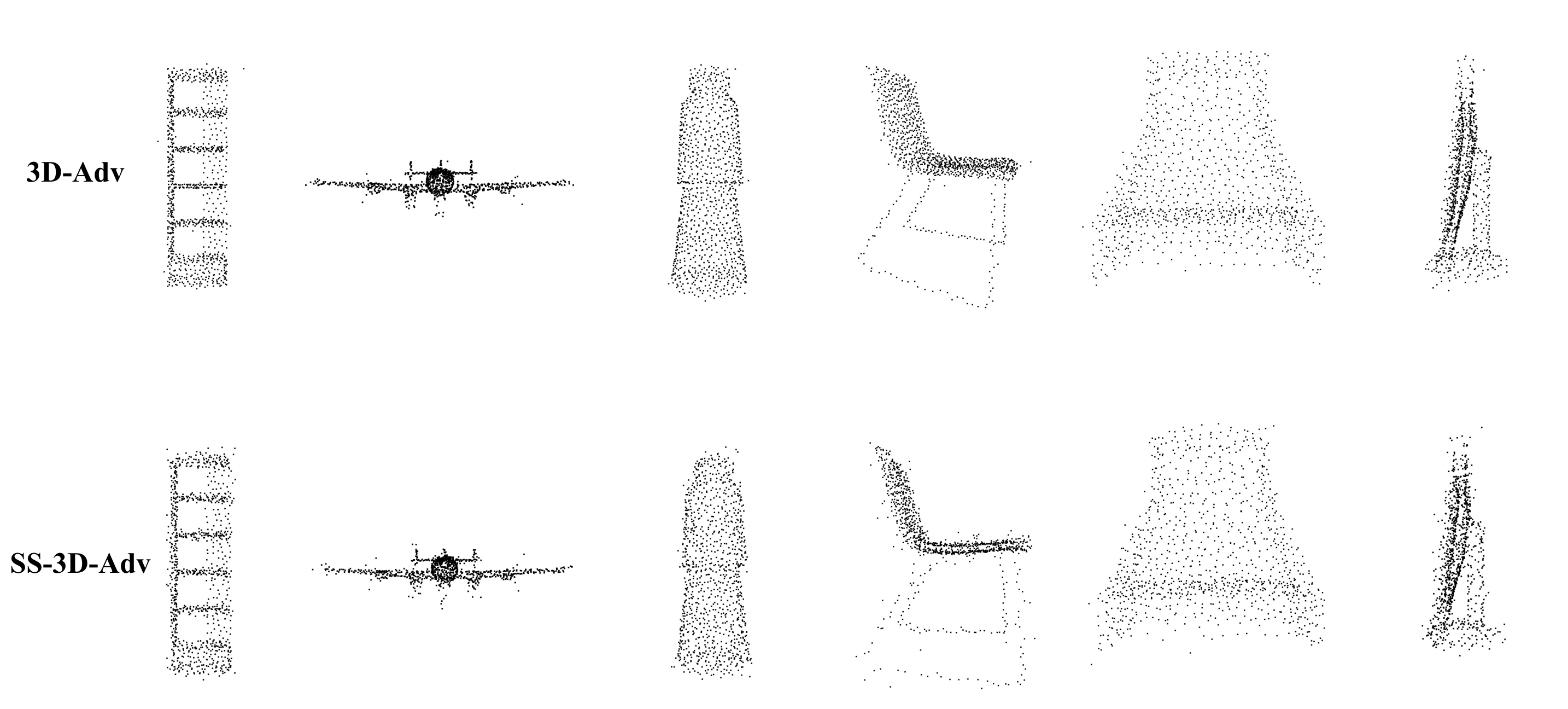}
    \caption{The comparison of adversarial point cloud samples generated by SS-3D-Adv and 3D-Adv attacks.}
    \label{ss_3dadv_compare}
\end{figure}

As shown in Figure \ref{ss_3dadv_compare}, for the SS-3D-Adv attack, the disturbance on the surface of the adversarial point cloud is more than that of 3D-Adv. This phenomenon can be clearly observed from the chair and the bottle. 

\begin{figure}[htp]
    \centering
    \includegraphics[width=\linewidth]{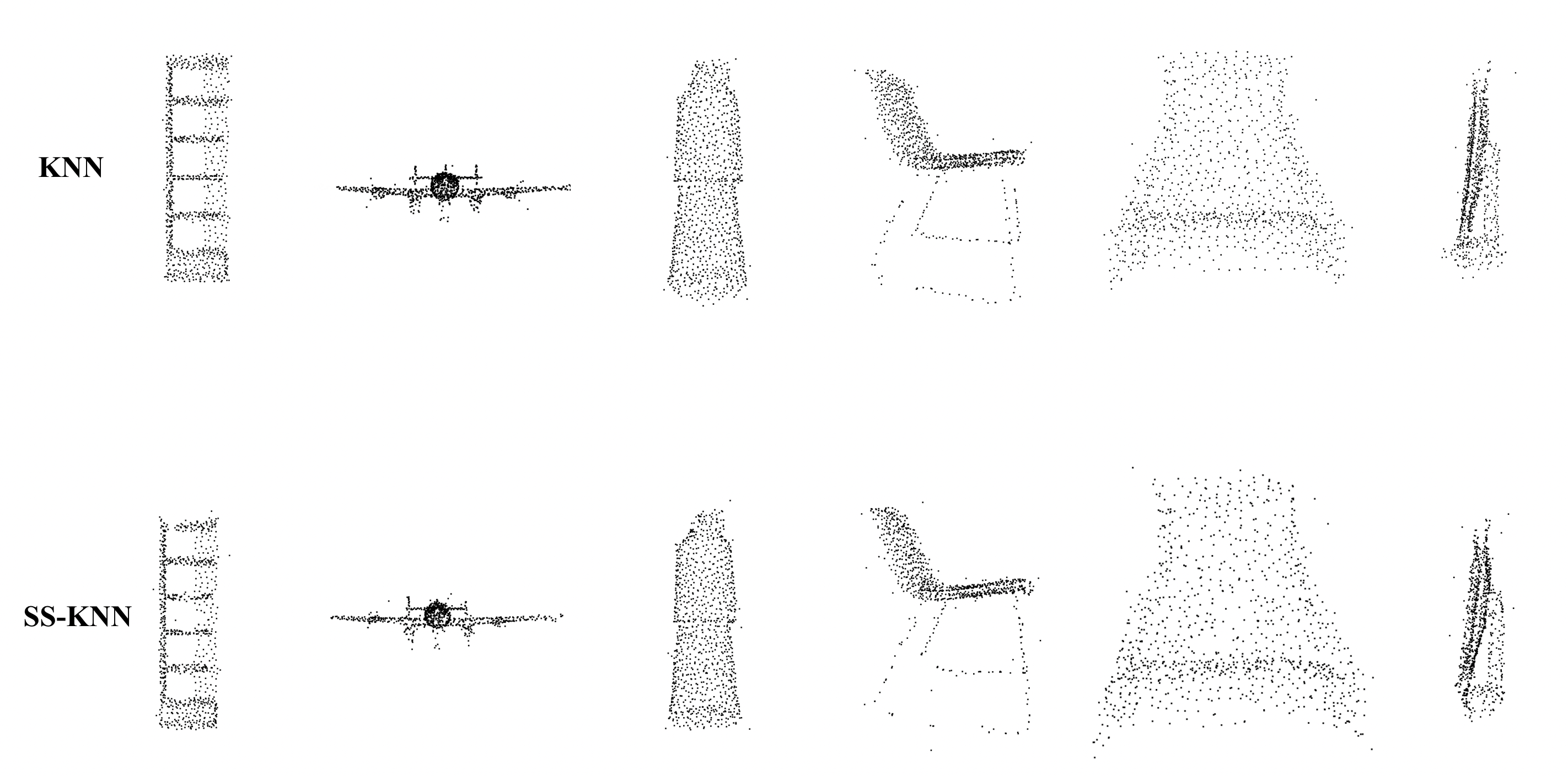}
    \caption{The comparison of adversarial point cloud samples generated by SS-KNN and KNN attacks.}
    \label{ss_knn_compare}
\end{figure}

As shown in Figure \ref{ss_knn_compare}, it can be observed from the monitor object that for SS-KNN attack, it produces more perturbations at the surface of the adversarial point cloud than KNN. A similar phenomenon can be observed for the object of the airplane.

\begin{figure}[htp]
    \centering
    \includegraphics[width=\linewidth]{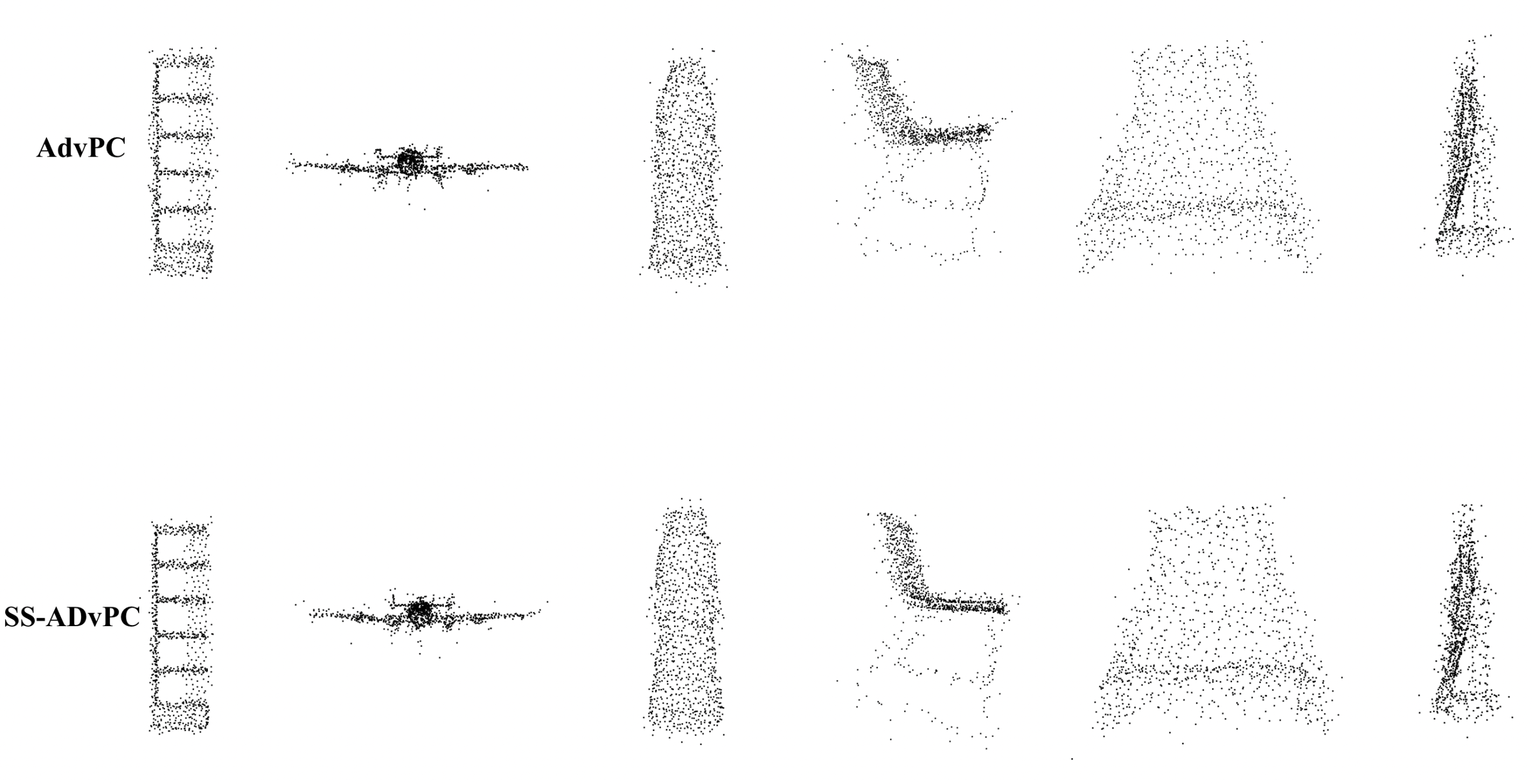}
    \caption{The comparison of adversarial point cloud samples generated by SS-AdvPC and AdvPC attacks.}
    \label{ss_advpc_compare}
\end{figure}

As shown in Figure \ref{ss_advpc_compare}, for SS-AdvPC attack, it generates significantly more noise points than AdvPC, but most of these noise points still fit the surface of the object, so SS-AdvPC will be more difficult defend than AdvPC.

\begin{figure}[htp]
    \centering
    \includegraphics[width=\linewidth]{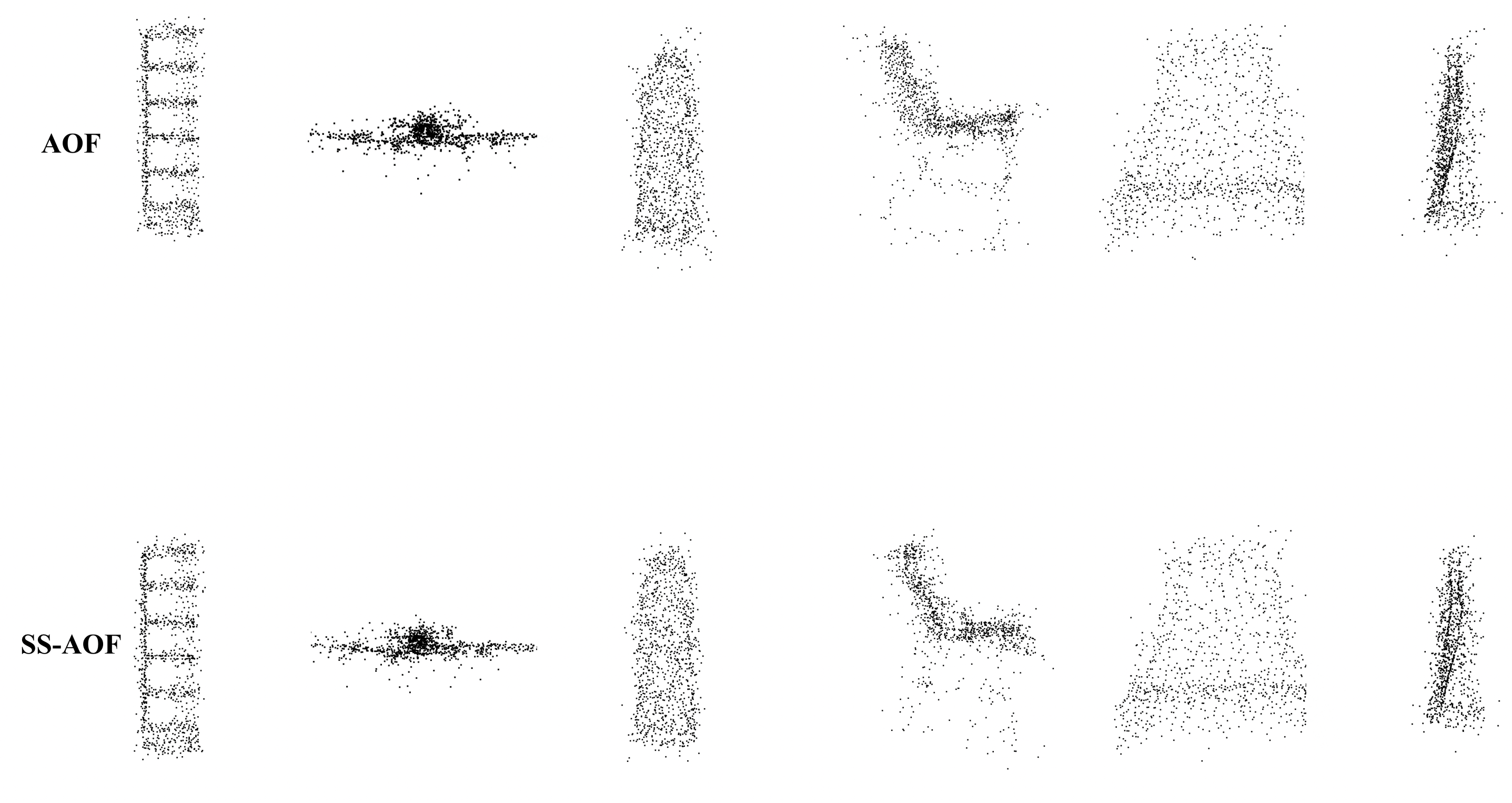}
    \caption{The comparison of adversarial point cloud samples generated by SS-AOF and AOF attacks.}
    \label{ss_aof_compare}
\end{figure}

The comparison of adversarial samples generated by SS-AOF and AOF attacks is shown in Figure \ref{ss_aof_compare}. For AOF attacks, the noise points are obviously more than the previous attacks, but SS-AOF can effectively reduces outliers that are too far away from surface, this phenomenon can be observed from two objects such as airplanes and bottles. Most of these noise points generated by SS-AOF are still attached to the surface of the object, so SS-AOF will be harder to defend than AOF.

\subsection{The transferability of targeted attack.}
This section implements targeted attacks and provides a fair comparison of their transferability performance. For targeted attacks, we evaluates various attack algorithms using the classification accuracy and targeted attack success rate. Table \ref{targettransfer} shows the results of the targeted attack success rate when the adversarial examples generated by SS-KNN and KNN on PointNet are transferred to other black-box models. When adversarial examples are transferred to other black-box models, the higher the targeted attack success rate of adversarial examples, the better the transferability. It can be seen from the results in the table \ref{targettransfer} that the transferability performance of the targeted attack is relatively low compared to the untargeted attack's transferability, which indicate that the targeted attack is difficult to transfer.

As shown in the table \ref{targettransfer}, when the targeted attack of SS-KNN and KNN is transferred to other black-box models, the targeted attack success rate of SS-KNN relative to KNN is basically not more than 2\%. However, the SS-KNN attack has higher targeted attack success rate than KNN attack in most cases, which further illustrates the effectiveness of the SS attack. When the targeted attack success rate of the SS-KNN attack method on PointNet++ is 0.77\%, PointConv is 1.38\%, and DGCNN is 1.34\%, while the baseline attack method KNN attack only obtains the targeted attack success rate of 0.65\%, 1.18\% and 1.18\%, respectively. In addition, except for a few attacks that are slightly lower than the baseline attack methods when transfered to other black-box models, the other attack methods combined with the SS attack greatly outperform the baseline methods, which further strongly proves that and the combination of SS attack and mainstream 3D point cloud attack methods is effective in improving the targeted attack success rate of 3D adversarial point cloud samples.

\begin{table}[ht]
\centering
\caption{\textbf{Targeted attack success rate of Attacks for SS-KNN and KNN.} 
}
\setlength{\tabcolsep}{1.7mm} %
\renewcommand{\arraystretch}{1.1} %
\resizebox{0.9\linewidth}{!}{
\begin{tabular}{cc|cccc} 
\toprule
%  &  &\multicolumn{4}{c}{$\epsilon_\infty = 0.18$}\\
Victim models & Attacks  & PointNet & PointNet++ & PointConv & DGCNN   \\
\midrule

 \multirow{2}*{PointNet}& KNN  & 100 &  0.65 & 1.18 &  1.18  \\
  & SS-KNN  & 100 &  \bf0.77  \textcolor{green!40!gray}{\small $\uparrow$0.12}& \bf1.38  \textcolor{green!40!gray}{\small $\uparrow$0.2}&  \bf1.34   \textcolor{green!40!gray}{\small $\uparrow$0.16}\\
 \midrule

 \multirow{2}*{PointNet++}& KNN  & 0.53 &  100 & 1.30 &  1.13  \\
  & SS-KNN  & \bf0.65 \textcolor{green!40!gray}{\small $\uparrow$0.12}& 100  & 1.09 &  \bf1.34 \textcolor{green!40!gray}{\small $\uparrow$0.21}\\
 \midrule
  
 \multirow{2}*{PointConv}& KNN  & 0.69 &  3.57 & 89.1 &  2.47  \\
  & SS-KNN  & \bf0.77 \textcolor{green!40!gray}{\small $\uparrow$0.08} &  \bf3.65 \textcolor{green!40!gray}{\small $\uparrow$0.08} & 89.2 &  2.39  \\
   \midrule
  \multirow{2}*{DGCNN}& KNN  & 0.97 &  2.27 & 2.03 &  59.9  \\
  & SS-KNN  & \bf1.26 \textcolor{green!40!gray}{\small $\uparrow$0.29} &  \bf2.47 \textcolor{green!40!gray}{\small $\uparrow$0.2} & 1.90 &  60.9\\
 \bottomrule
\end{tabular}}
\label{targettransfer}
\end{table}

Table \ref{acctransfer} shows the classification accuracy results when the adversarial examples generated by SS-KNN and KNN on PointNet are transferred to other black-box models. When adversarial examples are transferred to other black-box models, the higher the classification accuracy of adversarial examples, the worse the transferability. For the adversarial point cloud samples generated on PointNet, the classification accuracy of the SS-KNN attack method on PointNet++ is 79.05\%, DGCNN is 75.41\%, and PointConv is 78.53\%, while AOF attack achieves classification accuracys of 80.51\%, 76.66\% and 80.51\%, respectively. It can be seen that the classification accuracys of the SS-KNN attack methods are significantly lower than the baseline methods, which further strongly proves that the combination of SS attacks and the current mainstream 3D point cloud attack methods is effective in improving the transferability of 3D adversarial point cloud samples.

\begin{table}[htp]
\centering
\caption{\textbf{Accuracy of Targeted Attacks for SS-KNN and KNN.} 
}
\setlength{\tabcolsep}{1.7mm} %
\renewcommand{\arraystretch}{1.1} %
\resizebox{0.9\linewidth}{!}{
\begin{tabular}{cc|cccc} 
\toprule
%  &  &\multicolumn{4}{c}{$\epsilon_\infty = 0.18$}\\
Victim models & Attacks  & PointNet & PointNet++ & PointConv & DGCNN   \\
\midrule
 \multirow{2}*{PointNet}& KNN  & 0 & 80.51 &  76.66 & 80.51  \\
  & SS-KNN  & 0 &  \bf79.05 \textcolor{green!40!gray}{\small $\downarrow$1.46} & \bf75.41 \textcolor{green!40!gray}{\small $\downarrow$1.25}&  \bf78.53 \textcolor{green!40!gray}{\small $\downarrow$1.98} \\
 \midrule
 \multirow{2}*{PointNet++}& KNN  & 84.16 &  0 & 79.66 &  82.13  \\
  & SS-KNN  & \bf83.51  \textcolor{green!40!gray}{\small $\downarrow$0.65}&  0 & \bf78.00  \textcolor{green!40!gray}{\small $\downarrow$1.66} &  \bf81.60  \textcolor{green!40!gray}{\small $\downarrow$0.53} \\
 \midrule
  
 \multirow{2}*{PointConv}& KNN  & 81.52 & 71.80 & 3.16 &  76.09  \\
  & SS-KNN  & 81.60 &  \bf71.60 \textcolor{green!40!gray}{\small $\downarrow$0.20} & \bf2.80 \textcolor{green!40!gray}{\small $\downarrow$0.36}&  \bf75.81 \textcolor{green!40!gray}{\small $\downarrow$0.28} \\
   \midrule
  \multirow{2}*{DGCNN}& KNN  & 79.34 &  68.44 & 65.92 &  19.77  \\
  & SS-KNN  & 79.34 &  \bf66.57 \textcolor{green!40!gray}{\small $\downarrow$1.87} & \bf65.68 \textcolor{green!40!gray}{\small $\downarrow$2.76}&  \bf18.56 \textcolor{green!40!gray}{\small $\downarrow$1.21} \\
 \bottomrule
\end{tabular}}
\label{acctransfer}
\end{table}

\subsection{Performance under defense}

We further tests the performance under various defense. The defense performances under no defense, and under the Simple Random Sampling~\cite{yang2019adversarial} (SRS), Statistical Outlier Removal~\cite{yu2018pu} (SOR), DUP-Net~\cite{dupnet} and the IF-Defense~\cite{ifdefense} methods are compared. Since this paper mainly focuses on the transferability of 3D adversarial point cloud examples, the experiments in this section will mainly focus on the transferability after defense. The defense algorithms in this paper are all implemented by IF-Defense's open sourced code. Among them, ConvNet-Opt, Onet-Remesh, and ONet-Opt are three variants of IF-Defense.

From the results in Table \ref{3D-Adv_p1_underdefense}, we can observed that the transferability of SS attack under most defenses is better than the baseline attacks. Specifically, when the adversarial samples craft by the SS-3D-Adv attack on PointNet under SRS defense are transferred to PointNet++, PointConv and DGCNN, the transferability of the 3D-Adv attack on these models is improved 12.90\%, 5.08\%, 4.39\% respectively compared with the 3D-Adv attack, the similar situation can be observed for other defenses. These results further demonstrate the effectiveness of the SS attack methods proposed in this paper.

\begin{table}[htp]
\centering
\caption{\textbf{Transferability of various Attacks on PointNet under various defenses.} We use $T_{rans}$ defined in Eq (\ref{eqn:trans}) to evaluate the transferability. 
}
\setlength{\tabcolsep}{1.7mm} %
\renewcommand{\arraystretch}{1.1} %
\resizebox{0.9\linewidth}{!}{
\begin{tabular}{cc|cccc} 
\toprule
%  &  &\multicolumn{4}{c}{$\epsilon_\infty = 0.18$}\\
Defenses & Attacks  & PointNet & PointNet++ & PointConv & DGCNN   \\
\midrule

 \multirow{2}*{SRS}& 3D-Adv  & 36.93 &  33.79 & 14.53 &  20.69  \\
  & SS-3D-Adv  & \bf42.37 &  \bf46.69 & \bf19.61 &  \bf25.08   \\
 \midrule
  \multirow{2}*{SOR}& 3D-Adv  & 18.33 & 4.06 & 3.22 &  5.65  \\
  & SS-3D-Adv  & \bf23.91 &  \bf7.13 & \bf4.38 &  \bf6.27   \\
 \midrule
 \multirow{2}*{DUP-Net}& 3D-Adv  & 9.89 & 5.89 & 7.75 & 8.69  \\
  & SS-3D-Adv  & \bf11.66 &  \bf7.90 & \bf7.88 &  \bf9.01   \\
 \midrule

 \multirow{2}*{ConvNet-Opt}& 3D-Adv  & 6.17 &  11.16 & 9.62 &  11.02  \\
  & SS-3D-Adv  & \bf6.26 &  \bf11.29 & 9.31 &  \bf11.20   \\
 \midrule
 \multirow{2}*{Onet-Remesh}& 3D-Adv  & 11.52 & 24.19 & 20.05 &  24.55  \\
  & SS-3D-Adv  & \bf11.66 &  \bf25.44 & \bf22.14 & 24.46  \\
 \midrule
 \multirow{2}*{ONet-Opt}& 3D-Adv  & 6.62 &  13.92 & 13.05 &  13.04  \\
  & SS-3D-Adv  & 6.26 &  \bf14.73 & \bf13.36 &  \bf12.81   \\

 \midrule

 \multirow{2}*{SRS}& KNN  & 91.65 &  40.80 & 18.36 &  25.71  \\
  & SS-KNN  & \bf96.64 &  \bf51.69 & \bf24.28 &  \bf31.45   \\
 \midrule
  \multirow{2}*{SOR}& KNN  &76.45 & 6.86 & 4.69 & 7.52  \\
  & SS-KNN  & 73.41 &  \bf9.32 & \bf5.18 &  \bf9.54   \\
 \midrule
 \multirow{2}*{DUP-Net}& KNN  & 29.90 & 8.12 & 7.39 & 11.33  \\
  & SS-KNN  & \bf47.32& \bf9.82  & \bf8.64 &  \bf11.74 \\
 \midrule

 \multirow{2}*{ConvNet-Opt}& KNN  & 11.11 & 10.93 & 10.26 &  11.51  \\
  & SS-KNN  & \bf17.01 &  \bf12.14 & 9.62 &  11.51   \\
 \midrule
 \multirow{2}*{Onet-Remesh}& KNN  & 13.06 & 24.95 & 21.56 &  25.17  \\
  & SS-KNN  & \bf14.01 &  \bf25.49 & \bf21.79 & \bf26.97  \\
 \midrule
 \multirow{2}*{ONet-Opt}& KNN  & 9.25 &  14.33 & 12.87 &  13.35  \\
  & SS-KNN  & \bf11.66 &  \bf15.71 & \bf13.50 &  \bf14.25   \\

\midrule

 \multirow{2}*{SRS}& AdvPC  & 89.38 &  63.30 & 28.96 &  33.56  \\
  & SS-AdvPC  & 82.53 &  \bf64.46 & \bf30.43 &  \bf35.35   \\
 \midrule
  \multirow{2}*{SOR}& AdvPC  & 53.22 &  16.32 & 6.39 &  10.52  \\
  & SS-AdvPC  & 42.60 &  \bf17.35 & 5.99 &  10.12   \\
 \midrule
 \multirow{2}*{DUP-Net}& AdvPC  & 23.77 &  16.83 & 12.03 &  14.65  \\
  & SS-AdvPC  & 21.14& \bf17.32  & 9.98 &  \bf14.82 \\

 \midrule

 \multirow{2}*{ConvNet-Opt}& AdvPC  & 9.94 &  13.84 & 11.41 &  12.54  \\
  & SS-AdvPC  & 8.75 &  13.57 & \bf11.54 &  12.09   \\
 \midrule
 \multirow{2}*{Onet-Remesh}& AdvPC  & 12.65 & 27.36 & 23.75 &  26.39  \\
  & SS-AdvPC  & 12.11 &  \bf30.35 & \bf24.19 &  \bf27.01   \\
 \midrule
 \multirow{2}*{ONet-Opt}& AdvPC  & 7.95 &  17.90 & 15.06 &  13.58  \\
  & SS-AdvPC  & \bf8.66 &  17.36 & \bf15.33 &  \bf15.32   \\
 
 \midrule

 \multirow{2}*{SRS}& AOF  & 97.77 &  77.90 & 49.06 &  49.99  \\
  & SS-AOF  & 96.55 &  \bf78.30 & \bf50.89 &  49.91   \\
 \midrule
  \multirow{2}*{SOR}& AOF  & 90.83 &  41.88 & 21.33 &  24.86  \\
  & SS-AOF  & 85.79 &  \bf43.17 & \bf22.09 &  24.55   \\
 \midrule
 \multirow{2}*{DUP-Net}& AOF  & 64.65 &  38.66 & 26.20 &  28.40  \\
  & SS-AOF  & 60.52& \bf41.78  & \bf27.40 &  27.68 \\
 \midrule

 \multirow{2}*{ConvNet-Opt}& AOF  & 20.19 &  27.05 & 22.10 &  18.90  \\
  & SS-AOF  & 19.37 &  \bf27.76 & \bf25.04 &  \bf19.17   \\
 \midrule
 \multirow{2}*{Onet-Remesh}& AOF  & 17.60 & 37.27 & 31.50 &  33.91  \\
  & SS-AOF  & 17.37 &  37.05 & \bf31.86 & 33.33  \\
 \midrule
 \multirow{2}*{ONet-Opt}& AOF  & 14.33 &  26.47 & 24.28 &  17.92  \\
  & SS-AOF  & 12.38 &  26.07 & \bf26.29 &  \bf20.16   \\

 \bottomrule
\end{tabular}}
\label{3D-Adv_p1_underdefense}
\end{table}

\subsection{Comparative experiments}
In this section, we conducts a series of comparative experiments to analyze the impact of various parameters on the SS attacks.

\subsubsection{Influence of $p_a$}
\begin{figure}[thp]
    \centering
    \includegraphics[width=0.7\linewidth]{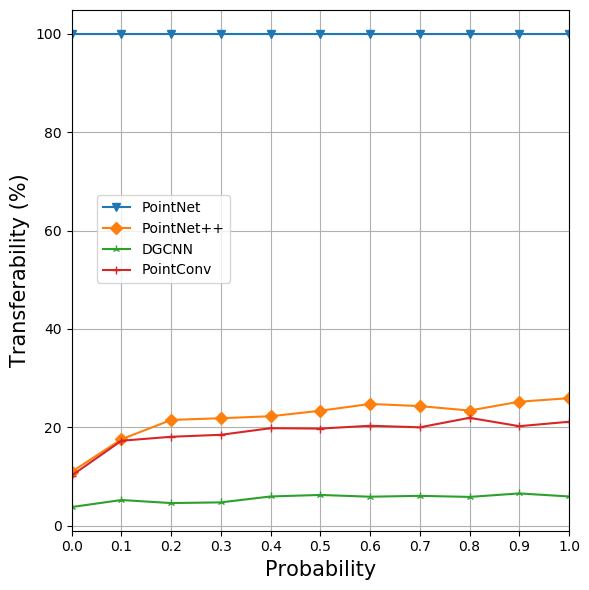}
    \caption{The effect of $p_a$ on transferability.}
    \label{pa_relations}
\end{figure}

We use SS-KNN for experiments, and set $p_s=0.5$. As shown in Figure \ref{pa_relations}, it can be seen that when $p_a$ is only 0.1, the transferability of the 3D adversarial point cloud samples generated on PointNet to PointNet++, DGCNN, and PointConv is greatly improved. This further validates the effect of transforming the input point cloud to mitigate overfitting and improve transferability. When $p_a$ slowly increases, the transferability will further improve, until $p_a=0.8$ when there is a slow decline. Therefore, in order to ensure the higher transferability, this paper takes $p_a=0.7$.

\subsubsection{Influence of $p_s$}
\begin{figure}[thp]
    \centering
    \includegraphics[width=0.7\linewidth]{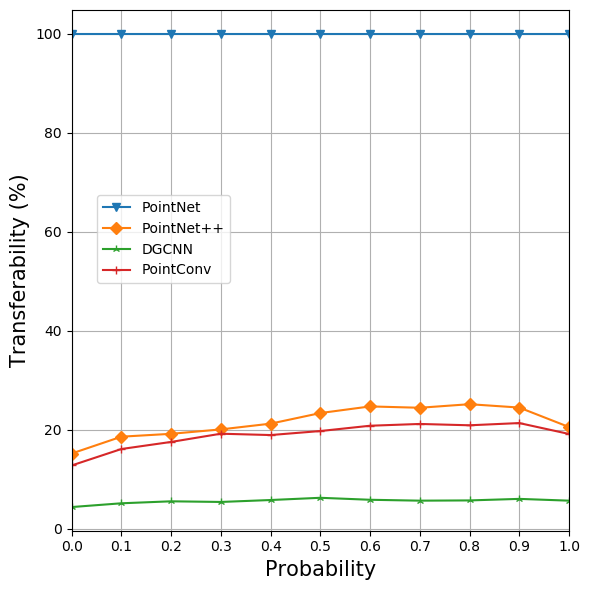}
    \caption{The effect of $p_s$ on transferability.}
    \label{ps_relations}
\end{figure}

We set $p_a=0.5$ here. As shown in Figure \ref{ps_relations}, it can be seen that when $p_s$ is only 0.1, the transferability of the 3D adversarial point cloud samples generated on PointNet to PointNet++, DGCNN, and PointConv is improved siginificantly. This further verifies that shear the input point cloud can improve transferability. When $p_s$ is slowly increased, the transferability will be further improved, until $p_a=0.9$ to $p_a=1.0$, the transferability has dropped significantly and rapidly. Therefore, in order to ensure the higher transferability, this paper takes $p_s=0.7$.

\subsubsection{Influence of attack iterations}
\begin{figure}[thp]
    \centering
    \includegraphics[width=0.7\linewidth]{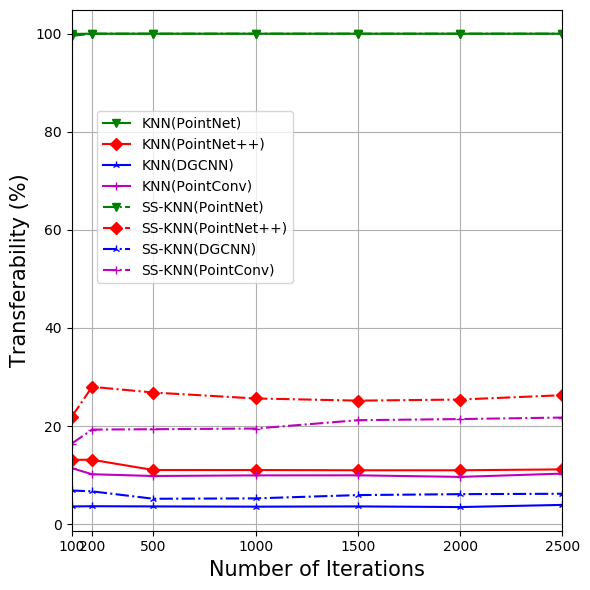}
    \caption{The effect of attack iterations.}
    \label{numiter_relations}
\end{figure}

In order to further explore the effect of the number of iterations on the transferability of the SS attack, we use PointNet as the white-box model, and conducts experiments on the iterations of $\{100, 200, 500, 1000, 1500, 2000, 2500\}$ respectively, and the experimental results are shown in Figure \ref{numiter_relations}. As can be seen from the figure, the transferability of the SS-KNN attack method to the black-box models PointNet++, DGCNN, and PointConv under different iterations is greatly improved compared with the KNN attack method, and the impact of different iterations on the number of iterations is limited, and the transferability is almost constant between 1000 and 2500 iterations.

\subsubsection{Influence of attack budgets}
\begin{figure}[thp]
    \centering
    \includegraphics[width=0.7\linewidth]{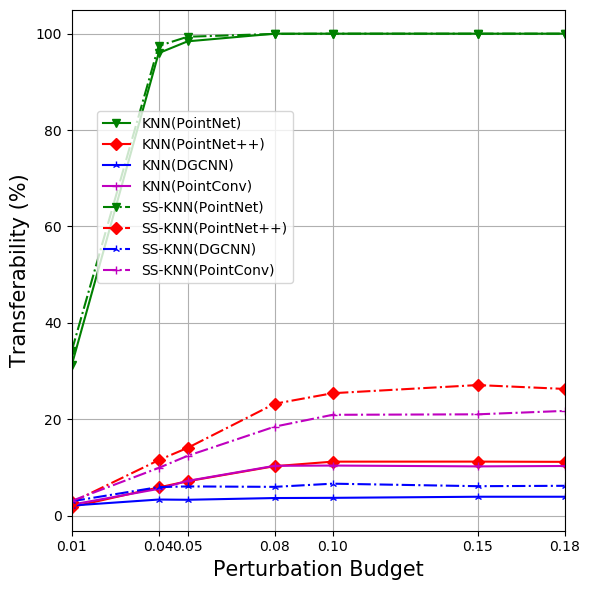}
    \caption{The effect of attack budgets.}
    \label{budget_relations}
\end{figure}

In order to further explore the impact of the attack budget on the transferability of the attack, we use PointNet as the white-box model, and conducts experiments on the attack budgets of $\{0.01, 0.04, 0.05, 0.08, 0.10, 0.15, 0.18\}$ respectively. The experimental results are shown in Figure \ref{budget_relations}, the transferability of the SS-KNN attack method to the black-box models PointNet++, DGCNN, and PointConv under different attack budgets is greatly improved compared to the KNN attack method. When the attack budgets is between 0.01 and 0.01, the transferability of the SS-KNN attack method increases rapidly, while the transferability of the KNN attack method remains almost unchanged.

\section{Conclusion}\label{sec:Conclusion}

In this paper, we propose Scale and Shear Attack (SS attack), a novel and plug-and-play attack method for 3D point clouds. It reduces the overfitting of the adversarial attack on the victim model by randomly scale and shear the input 3D point cloud, thereby improving the transferability of the adversarial samples. Extensive experiments show that the method proposed in this paper can improve the transferability of untargeted attacks significantly. In addition, for the challenging targeted attack, the SS attack can further boost the transferability, illustrating the effectiveness of the proposed SS attack. Moreover, SS attack can surpass baseline attack methods by a large margin under various defenses.

The dangerous targeted attacks is much greater than that of untargeted attacks, but current targeted attacks are have low transferability. Therefore, a promising research direction in the future is to improve the transferability of 3D point cloud targeted attacks.

% \section{Acknowledgments}
% The project was supported by Innovation Project of Guangxi Graduate Education (YCBZ2021019).

% \bibliographystyle{IEEETran.cls}
\bibliography{ref/ref}

\begin{thebibliography}{10}

\bibitem{lecun2015deepnature}
Y.~LeCun, Y.~Bengio, and G.~Hinton, ``Deep learning,'' {\em nature}, vol.~521,
  no.~7553, pp.~436--444, 2015.

\bibitem{benchmarkingimage}
Y.~Dong, Q.-A. Fu, X.~Yang, T.~Pang, H.~Su, Z.~Xiao, and J.~Zhu, ``Benchmarking
  adversarial robustness on image classification,'' in {\em Proceedings of the
  IEEE/CVF Conference on Computer Vision and Pattern Recognition},
  pp.~321--331, 2020.

\bibitem{szegedy2014intriguing}
C.~Szegedy, W.~Zaremba, I.~Sutskever, J.~Bruna, D.~Erhan, I.~Goodfellow, and
  R.~Fergus, ``Intriguing properties of neural networks,'' in {\em 2nd
  International Conference on Learning Representations, ICLR 2014}, 2014.

\bibitem{adsurvey}
N.~Akhtar and A.~Mian, ``Threat of adversarial attacks on deep learning in
  computer vision: A survey,'' {\em Ieee Access}, vol.~6, pp.~14410--14430,
  2018.

\bibitem{knnattack}
T.~Tsai, K.~Yang, T.-Y. Ho, and Y.~Jin, ``Robust adversarial objects against
  deep learning models,'' in {\em Proceedings of the AAAI Conference on
  Artificial Intelligence}, vol.~34, pp.~954--962, 2020.

\bibitem{advpc}
A.~Hamdi, S.~Rojas, A.~Thabet, and B.~Ghanem, ``Advpc: Transferable adversarial
  perturbations on 3d point clouds,'' in {\em European Conference on Computer
  Vision}, pp.~241--257, Springer, 2020.

\bibitem{aof_attack}
B.~Liu, J.~Zhang, and J.~Zhu, ``Boosting 3d adversarial attacks with attacking
  on frequency,'' {\em IEEE Access}, vol.~10, pp.~50974--50984, 2022.

\bibitem{cwattack}
N.~Carlini and D.~Wagner, ``Towards evaluating the robustness of neural
  networks,'' in {\em 2017 ieee symposium on security and privacy (sp)},
  pp.~39--57, IEEE, 2017.

\bibitem{cutmix}
S.~Yun, D.~Han, S.~J. Oh, S.~Chun, J.~Choe, and Y.~Yoo, ``Cutmix:
  Regularization strategy to train strong classifiers with localizable
  features,'' in {\em Proceedings of the IEEE International Conference on
  Computer Vision}, pp.~6023--6032, 2019.

\bibitem{generatingadpoint}
C.~Xiang, C.~R. Qi, and B.~Li, ``Generating 3d adversarial point clouds,'' in
  {\em Proceedings of the IEEE/CVF Conference on Computer Vision and Pattern
  Recognition}, pp.~9136--9144, 2019.

\bibitem{pointnet}
C.~R. Qi, H.~Su, K.~Mo, and L.~J. Guibas, ``Pointnet: Deep learning on point
  sets for 3d classification and segmentation,'' in {\em Proceedings of the
  IEEE conference on computer vision and pattern recognition}, pp.~652--660,
  2017.

\bibitem{pointnet++}
C.~R. Qi, L.~Yi, H.~Su, and L.~J. Guibas, ``Pointnet++: Deep hierarchical
  feature learning on point sets in a metric space,'' {\em Advances in Neural
  Information Processing Systems}, vol.~30, 2017.

\bibitem{pointconv}
W.~Wu, Z.~Qi, and L.~Fuxin, ``Pointconv: Deep convolutional networks on 3d
  point clouds,'' in {\em Proceedings of the IEEE/CVF Conference on Computer
  Vision and Pattern Recognition}, pp.~9621--9630, 2019.

\bibitem{dgcnn}
Y.~Wang, Y.~Sun, Z.~Liu, S.~E. Sarma, M.~M. Bronstein, and J.~M. Solomon,
  ``Dynamic graph cnn for learning on point clouds,'' {\em Acm Transactions On
  Graphics (tog)}, vol.~38, no.~5, pp.~1--12, 2019.

\bibitem{guo2021pct}
M.-H. Guo, J.-X. Cai, Z.-N. Liu, T.-J. Mu, R.~R. Martin, and S.-M. Hu, ``Pct:
  Point cloud transformer,'' {\em Computational Visual Media}, vol.~7, no.~2,
  pp.~187--199, 2021.

\bibitem{han2022dual}
X.-F. Han, Y.-F. Jin, H.-X. Cheng, and G.-Q. Xiao, ``Dual transformer for point
  cloud analysis,'' {\em IEEE Transactions on Multimedia}, 2022.

\bibitem{zhang2022patchformer}
C.~Zhang, H.~Wan, X.~Shen, and Z.~Wu, ``Patchformer: An efficient point
  transformer with patch attention,'' in {\em Proceedings of the IEEE/CVF
  Conference on Computer Vision and Pattern Recognition}, pp.~11799--11808,
  2022.

\bibitem{geoa3}
Y.~Wen, J.~Lin, K.~Chen, C.~P. Chen, and K.~Jia, ``Geometry-aware generation of
  adversarial point clouds,'' {\em IEEE Transactions on Pattern Analysis and
  Machine Intelligence}, 2020.

\bibitem{pointcloudsaliencymaps}
T.~Zheng, C.~Chen, J.~Yuan, B.~Li, and K.~Ren, ``Pointcloud saliency maps,'' in
  {\em Proceedings of the IEEE/CVF International Conference on Computer
  Vision}, pp.~1598--1606, 2019.

\bibitem{dong2018boosting}
Y.~Dong, F.~Liao, T.~Pang, H.~Su, J.~Zhu, X.~Hu, and J.~Li, ``Boosting
  adversarial attacks with momentum,'' in {\em Proceedings of the IEEE
  conference on computer vision and pattern recognition}, pp.~9185--9193, 2018.

\bibitem{xie2019improving}
C.~Xie, Z.~Zhang, Y.~Zhou, S.~Bai, J.~Wang, Z.~Ren, and A.~L. Yuille,
  ``Improving transferability of adversarial examples with input diversity,''
  in {\em Proceedings of the IEEE/CVF Conference on Computer Vision and Pattern
  Recognition}, pp.~2730--2739, 2019.

\bibitem{dong2019evading}
Y.~Dong, T.~Pang, H.~Su, and J.~Zhu, ``Evading defenses to transferable
  adversarial examples by translation-invariant attacks,'' in {\em Proceedings
  of the IEEE/CVF Conference on Computer Vision and Pattern Recognition},
  pp.~4312--4321, 2019.

\bibitem{modelnet40}
Z.~Wu, S.~Song, A.~Khosla, F.~Yu, L.~Zhang, X.~Tang, and J.~Xiao, ``3d
  shapenets: A deep representation for volumetric shapes,'' in {\em Proceedings
  of the IEEE conference on computer vision and pattern recognition},
  pp.~1912--1920, 2015.

\bibitem{paszke2019pytorch}
A.~Paszke, S.~Gross, F.~Massa, A.~Lerer, J.~Bradbury, G.~Chanan, T.~Killeen,
  Z.~Lin, N.~Gimelshein, L.~Antiga, {\em et~al.}, ``Pytorch: An imperative
  style, high-performance deep learning library,'' {\em Advances in neural
  information processing systems}, vol.~32, pp.~8026--8037, 2019.

\bibitem{kingma2014adam}
D.~P. Kingma and J.~Ba, ``Adam: {A} method for stochastic optimization,'' in
  {\em 3rd International Conference on Learning Representations, {ICLR} 2015,
  San Diego, CA, USA, May 7-9, 2015, Conference Track Proceedings} (Y.~Bengio
  and Y.~LeCun, eds.), 2015.

\bibitem{yang2019adversarial}
J.~Yang, Q.~Zhang, R.~Fang, B.~Ni, J.~Liu, and Q.~Tian, ``Adversarial attack
  and defense on point sets,'' {\em arXiv preprint arXiv:1902.10899}, 2019.

\bibitem{yu2018pu}
L.~Yu, X.~Li, C.-W. Fu, D.~Cohen-Or, and P.-A. Heng, ``Pu-net: Point cloud
  upsampling network,'' in {\em Proceedings of the IEEE conference on computer
  vision and pattern recognition}, pp.~2790--2799, 2018.

\bibitem{dupnet}
H.~Zhou, K.~Chen, W.~Zhang, H.~Fang, W.~Zhou, and N.~Yu, ``Dup-net: Denoiser
  and upsampler network for 3d adversarial point clouds defense,'' in {\em
  Proceedings of the IEEE/CVF International Conference on Computer Vision},
  pp.~1961--1970, 2019.

\bibitem{ifdefense}
Z.~Wu, Y.~Duan, H.~Wang, Q.~Fan, and L.~J. Guibas, ``If-defense: 3d adversarial
  point cloud defense via implicit function based restoration,'' {\em CoRR},
  vol.~abs/2010.05272, 2020.

\end{thebibliography}

\vspace{-30pt}
\begin{IEEEbiography}[{\includegraphics[width=1in,height=1.25in,clip,keepaspectratio]{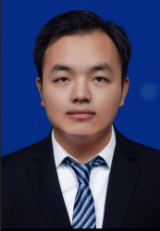}}]{Jinlai Zhang} received the B.S. degree from Changsha University of Science and Technology, Changsha, China, in 2017. He is currently pursuing the Ph.D. with the College of Mechanical Engineering, Guangxi University. His research interests include 3D deep learning and time series forecasting.
\end{IEEEbiography}
\vspace{-30pt}

\begin{IEEEbiography}[{\includegraphics[width=1in,height=1.25in,clip,keepaspectratio]{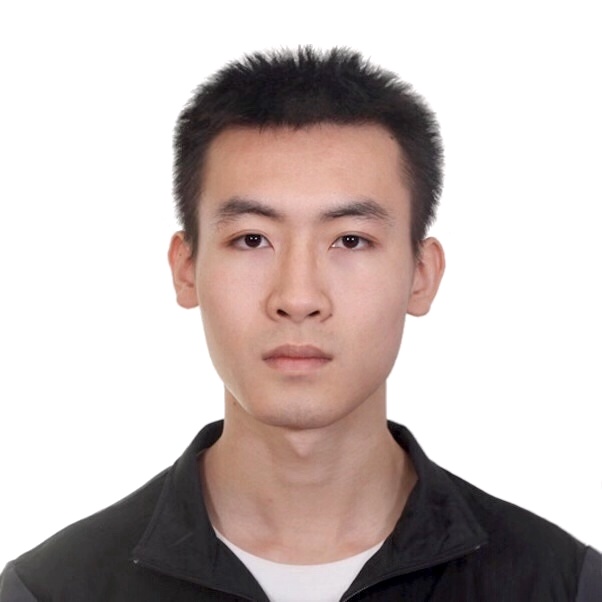}}]{Yinpeng Dong} received his BS and PhD degrees from the Department of Computer Science and Technology in Tsinghua University, where he is currently a postdoctoral researcher. His research interests are primarily on the adversarial robustness of machine learning and deep learning.
\end{IEEEbiography}
\vspace{-30pt}

\begin{IEEEbiography}[{\includegraphics[width=1in,height=1.25in,clip,keepaspectratio]{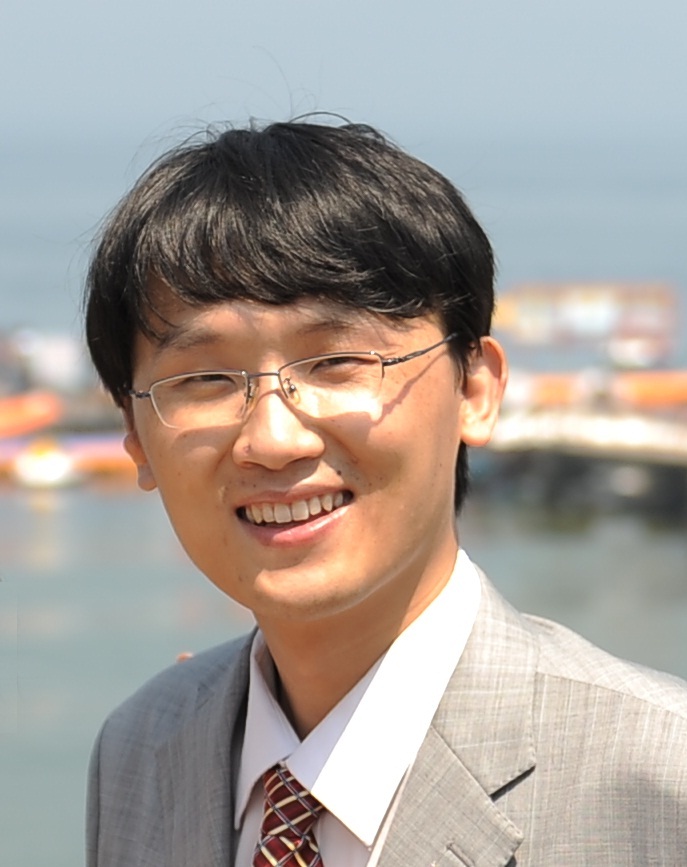}}]{Jun Zhu}  received his BS and PhD degrees from the Department of Computer Science and Technology in Tsinghua University, where he is currently a professor. He was an adjunct faculty and postdoctoral fellow in the Machine Learning Department, Carnegie Mellon University. His research interests are primarily on developing statistical machine learning methods to understand scientific and engineering data arising from various fields. He regularly serves as Area Chairs at prestigious conferences, including ICML, NeurIPS, ICLR, IJCAI and AAAI. He was selected as AI’s 10 to Watch by IEEE Intelligent Systems.
\end{IEEEbiography}
\vspace{-30pt}

% \begin{IEEEbiography}[{\includegraphics[width=1in,height=1.25in,clip,keepaspectratio]{photo/qizhi.png}}] {Qizhi Xie} received the BE. degree from the School of Information Engineering, China University of Geocience (Wuhan), in 2020. He is currently working towards the Ph.D. degree in the Department of Precision Instrument in Tsinghua University. His research interests include computer vision, deep learning, and deep reinforcement learning.
% \end{IEEEbiography}
% \vspace{-30pt}

\begin{IEEEbiography}[{\includegraphics[width=1in,height=1.25in,clip,keepaspectratio]{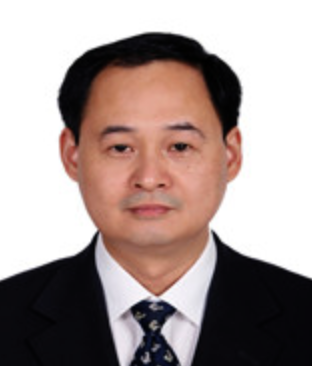}}]{Jihong Zhu}  received his Ph.D. degree from Nanjing University of Science and Technology, Nanjing, in 1995. Currently, he is a professor with the Department of Precision Instrument,Tsinghua University, Beijing, China. His research interests include flight control and robotics.
\end{IEEEbiography}
\vspace{-30pt}

\begin{IEEEbiography}[{\includegraphics[width=1in,height=1.25in,clip,keepaspectratio]{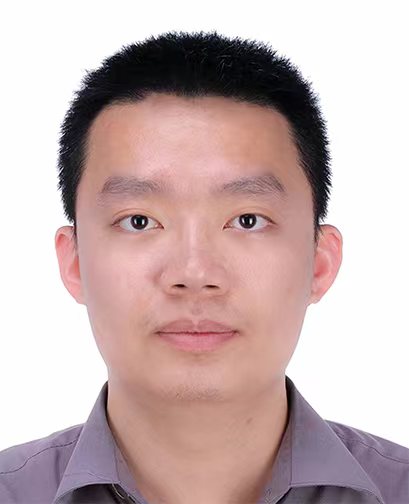}}]{Minchi Kuang}  received his doctorate from Tsinghua University in 2017. He is now working as an assistant researcher in the Department of precision instruments of Tsinghua University. His main research interests are artificial intelligence and intelligent control.
\end{IEEEbiography}
\vspace{-30pt}

\begin{IEEEbiography}[{\includegraphics[width=1in,height=1.25in,clip,keepaspectratio]{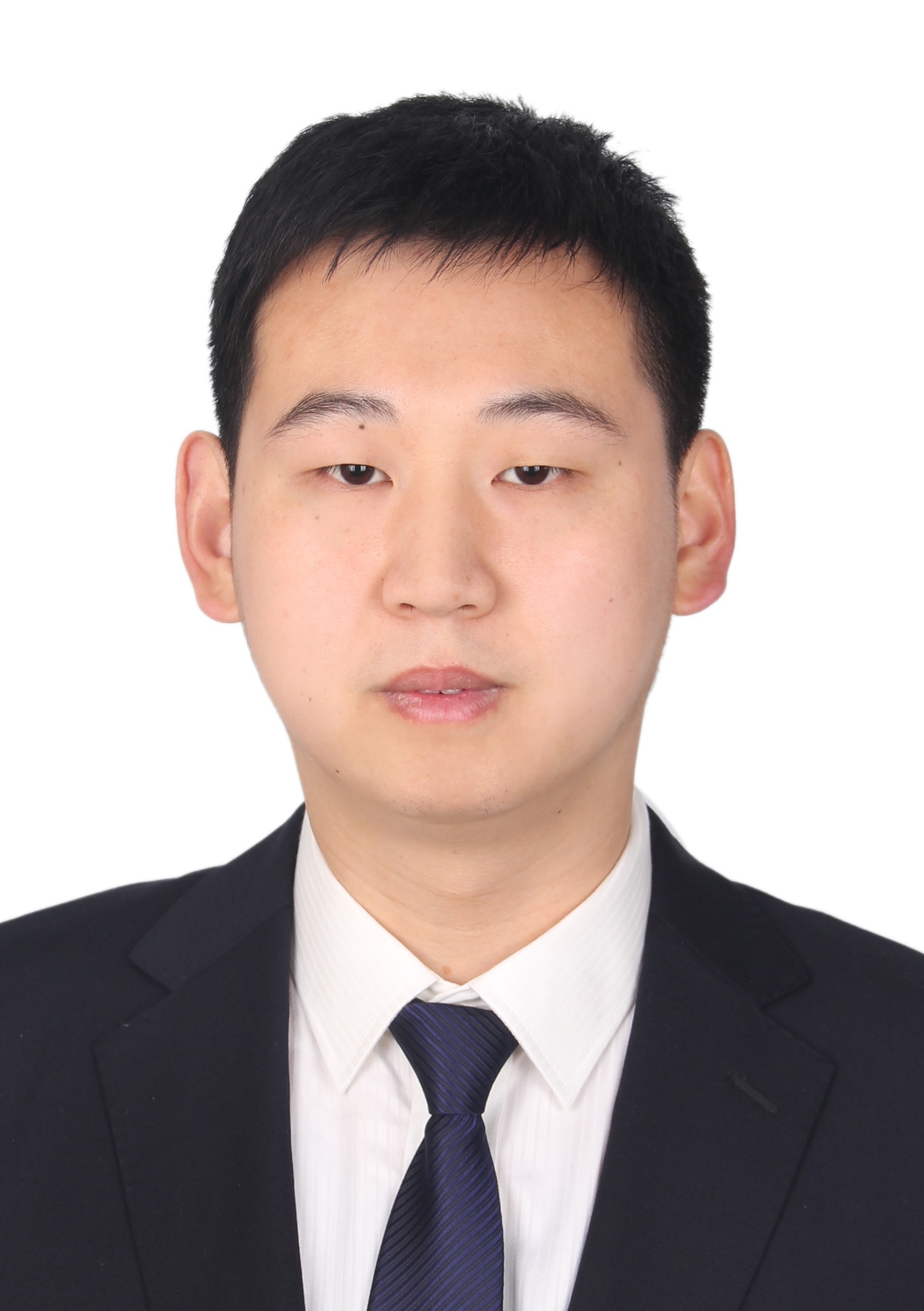}}]{Xiaming Yuan} received his Ph.D. degree from Tsinghua University, Beijing, in 2015. Currently, he is a research associate with the Department of Precision Instrument, Tsinghua University, Beijing, China. His research interests include flight control and information fusion.
\end{IEEEbiography}
\vspace{-30pt}

\end{document}